\documentclass{article}

\usepackage[final]{neurips_2023}

\usepackage[utf8]{inputenc} % allow utf-8 input
\usepackage[T1]{fontenc}    % use 8-bit T1 fonts
\usepackage{hyperref}       % hyperlinks
\usepackage{url}            % simple URL typesetting
\usepackage{booktabs}       % professional-quality tables
\usepackage{amsfonts}       % blackboard math symbols
\usepackage{nicefrac}       % compact symbols for 1/2, etc.
\usepackage{microtype}      % microtypography
\usepackage{xcolor}         % colors

\usepackage{amsmath}
\usepackage{amssymb}
\usepackage{mathtools}
\usepackage{amsthm}

\usepackage{microtype}
\usepackage{graphicx}
\usepackage{booktabs} % for professional tables

\usepackage{algorithmic}
\usepackage{algorithm}
\usepackage{bm}
\usepackage{wrapfig}
\usepackage{tikz}
\usepackage{scalerel}
\usepackage{xspace}
\usepackage{pifont}
\usepackage{multirow}
\usepackage{subcaption}
\theoremstyle{plain}

\theoremstyle{definition}

\theoremstyle{remark}

\newcommand{\val}[1]{{\fontfamily{pcr}\selectfont #1}}
\newcommand{\valsub}[2]{{\fontfamily{pcr}\selectfont #1}$_\text{#2}$}
\newcommand{\tikzcircle}[1][red,fill=red]{\scalerel*{\tikz \draw[#1] (0,0) circle (2pt);}{\circ}}

\newcommand{\tikzdiamond}[1][red,fill=red]{\scalerel*{\tikz \draw[rounded corners=0.06pt,#1] (-3pt,0)--++(45:3pt)--++(-45:3pt)--++(-180+45:3pt)--cycle;}{\diamond}}

\newcommand{\rigl}[1]{\tikzcircle[red,fill=red] \val{#1}\xspace }

\newcommand{\set}[1]{\tikzdiamond[blue, fill=blue] \val{#1}\xspace}

\newcommand\freefootnote[1]{%
  \let\thefootnote\relax%
  \footnotetext{#1}%
  \let\thefootnote\svthefootnote%
}
\usepackage[textsize=tiny]{todonotes}

\title{Balanced Training for Sparse GANs}

\author{Yite Wang$^{1,}  $\thanks{Equal contribution.}\,, Jing Wu$^{1,*}$, Naira Hovakimyan$^1$, Ruoyu Sun$^{2,3}$\thanks{Corresponding author.}  \\
$^1$University of Illinois Urbana-Champaign, USA\\
$^2$School of Data Science, The Chinese  University of Hong Kong, Shenzhen, China  \\
$^3$Shenzhen International Center for Industrial and Applied Mathematics,\\ Shenzhen Research Institute of Big Data\\ 
\texttt{\{yitew2,jingwu6,nhovakim\}@illinois.edu}, \texttt{sunruoyu@cuhk.edu.cn} \\
}

\begin{document}

\maketitle

\begin{abstract}
% In recent years, there has been a significant increase in the size of modern deep neural networks (DNNs), particularly for deep generative models like generative adversarial networks (GANs). Nevertheless, GANs tend to have high computational complexity, leading researchers to explore methods for reducing training and inference costs. One such method is dynamic sparse training (DST), which has gained popularity in the supervised learning field for its ability to maintain good performance while enjoying excellent training efficiency. Despite its potential, applying DST to GANs poses challenges due to the adversarial nature of the training process. In this paper, we propose a quantity called balance ratio (BR) to study the balance between the sparse generator and discriminator. We also propose a method called balanced dynamic sparse training (ADAPT), which aims to control BR during GAN training in order to achieve a good trade-off between performance and computational cost. Our code is available at \url{https://anonymous.4open.science/r/DST-GAN-207F/}.
Over the past few years, there has been growing interest in developing larger and deeper neural networks, including deep generative models like generative adversarial networks (GANs). However, GANs typically come with high computational complexity,  leading researchers to explore methods for reducing the training and inference costs.  One such approach gaining popularity in supervised learning is dynamic sparse training (DST), which maintains good performance while enjoying excellent training efficiency.  Despite its potential benefits, applying DST to GANs presents challenges due to the adversarial nature of the training process. In this paper, we propose a novel metric called the balance ratio (BR) to study the balance between the sparse generator and discriminator. We also introduce a new method called balanced dynamic sparse training (ADAPT), which seeks to control the BR during GAN training to achieve a good trade-off between performance and computational cost. Our proposed method shows promising results on multiple datasets, demonstrating its effectiveness. Our code is available at \url{https://github.com/YiteWang/ADAPT}.
% \freefootnote{$^\dag$Equal contribution.}
\end{abstract}

\section{Introduction}
\label{sect:intro}

Generative adversarial networks (GANs) \citep{goodfellow2020generative, biggan, sauer2022stylegan, lee2022autoregressive} are a type of generative model that has gained significant attention in recent years due to their impressive performance in image-generation tasks. However, the mainstream models in GANs are known to be computationally intensive, making them challenging to train in resource-constrained settings. Therefore, it is crucial to develop methods that can effectively reduce the computational cost of training GANs while maintaining their performance, making GANs more practical and applicable in real-world scenarios.

Neural network pruning has recently emerged as a powerful tool to reduce the training and inference costs of DNNs for supervised learning. There are mainly three genres of pruning methods, namely pruning-at-initialization, pruning-during-training, and post-hoc pruning methods. Post-hoc pruning \citep{janowsky1989pruning, OBD, han2015learning} can date back to the 1980s, which was first introduced for reducing inference time and memory requirements for efficient deployment; hence does not align with our purpose of efficient training. Later, pruning-at-initialization \citep{snip, grasp, synflow} and pruning-during-training methods \citep{wen2016learning} were introduced to circumvent the need to fully train the dense networks. However, early pruning-during-training algorithms \citep{louizos2017learning} do not bring much training efficiency compared to post-hoc pruning, while pruning-at-initialization methods usually suffer from significant performance drop \citep{frankle2021pruning}. Recently, advances in dynamic sparse training (DST) \citep{SET, evci2020rigging, liu2021sparse, liu2021selfish, ITOP} for the first time show that pruning-during-training methods can have comparable training FLOPs as pruning-at-initialization methods while having competing performance to post-hoc pruning. Therefore, applying DST on GANs seems to be a promising choice.

Although DST has attained remarkable achievements in supervised learning, the application of DST on GANs is not successful due to newly emerging challenges. One challenge is keeping the generator and the discriminator balanced. In particular, using overly strong discriminators can lead to overfitting, while weaker discriminators may fail to effectively prevent mode collapse \citep{arora2017generalization,bai2018approximability}. Hence, balancing the sparse generator and the (possibly) sparse discriminator throughout training is even more difficult. To mitigate the unbalance issue, a recent work STU-GAN \citep{liu2022don} proposes to apply DST directly to the generator. However, we find empirically that such an algorithm is likely to fail when the generator is already more powerful than the discriminator. Consequently, it remains unclear how to conduct balanced dynamic sparse training for GANs. 

To this end, we propose a metric called balance ratio (BR), which measures the degree of balance of the two components, to study sparse GAN training. We find that BR is useful in (1) understanding the interaction between the discriminator and the generator, (2) identifying the cause of a certain training failure/collapse \citep{biggan, chen2021data}, and (3) helping stabilize sparse GAN training as an indicator. To our best knowledge, this is the first study to quantify the unbalance of sparse GANs and may even provide new insights into dense GAN training.

Furthermore, using BR as an indicator, we propose b\underline{A}lanced \underline{D}yn\underline{A}mic s\underline{P}arse \underline{T}raining (\val{ADAPT}) to adjust the density and the connections of the discriminator automatically during training.

Our main contributions are summarized below:
\begin{itemize}
    \item We introduce a novel quantity named balance ratio to study the degree of balance in sparse GAN training.
    \item We find empirically that the balance ratio is problematic in certain practical training scenarios and that existing methods are inadequate for resolving this issue.
    \item We propose \val{ADAPT}, which makes real-time monitoring of the balance ratio. By dynamically adjusting the discriminator, \val{ADAPT} enables effective control of the balance ratio throughout training. Empirically, \val{ADAPT} achieves a good trade-off between performance and computational cost on several datasets.
\end{itemize}

\section{Related works}
\label{sect:related_works}

\subsection{Neural network pruning}
In deep learning, efficiency is achieved through several methods. This paper primarily focuses on model training and inference efficiency, which is different from techniques for data efficiency \citep{wang2018low, wu2023hallucination, wu2023genco}. These include neural architecture search (NAS) \citep{wang2022global, liu2018darts} to discover optimal network structures, quantization \citep{hubara2017quantized, rastegari2016xnor} for computational efficacy, knowledge distillation \citep{hinton2015distilling} to leverage the knowledge of larger models for smaller counterparts, and neural network pruning to remove unnecessary connections. Among these, neural network pruning is the focal point of our research. More specifically, we narrow our focus on unstructured pruning \citep{han2015learning, LTH}, 
 where individual weight is the finest resolution. This contracts with structured pruning \citep{liu2017learning, luo2017thinet, liu2018rethinking, huang2018data} where entire neurons or channels are pruned.  

% Based on the smallest granularity of pruned units, neural network pruning can be categorized into structured \citep{liu2017learning, luo2017thinet, liu2018rethinking, huang2018data} and unstructured pruning \citep{han2015learning, LTH}. We focus on unstructured pruning, where individual weight is the finest resolution. 

\textbf{Post-hoc pruning.} Post-hoc pruning method prunes weights of a fully-trained neural network. It usually requires high computational costs due to the multiple rounds of the train-prune-retrain procedure \citep{han2015learning, LTR}. Some use specific criteria \citep{OBD, OBS,han2015learning, guo2016dynamic, dong2017learning,   dai2018compressing, yu2018nisp, molchanov2019importance} to remove weights, while others perform extra optimization iterations \citep{verma2021sparsifying}. Post-hoc pruning was initially proposed to reduce the inference time, while lottery ticket works \citep{LTH, LTR} aimed to mine trainable sub-networks. 

\textbf{Pruning-at-initialization methods.} SNIP \citep{snip} is one of the pioneering works that aim to find trainable sub-networks without any training. Some follow-up works \citep{grasp, synflow, force, patil2021phew, prospr} aim to propose different metrics to prune networks at initialization. Among them, Synflow \citep{synflow}, SPP \citep{lee2019signal}, and FORCE \citep{force} try to address the problem of layer collapse during pruning. NTT \citep{NTT}, PHEW \citep{patil2021phew}, and NTK-SAP \citep{wang2022ntk} draw inspiration from neural tangent kernel theory. 

\textbf{Pruning-during-training methods.} Another genre of pruning algorithms prunes or adjusts DNNs throughout training. Early works add explicit $\ell_0$ \citep{louizos2017learning} or $\ell_1$ \citep{wen2016learning} regularization terms to encourage a sparse solution, hence mitigating performance drop incurred by post-hoc pruning. Later works learn the subnetworks structures through projected gradient descent \citep{zhou2021effective} or trainable masks \citep{srinivas2017training,xiao2019autoprune,kang2020operation, kusupati2020soft, liu2020dynamic, savarese2020winning}. However, these pruning-during-training methods often do not introduce memory sparsity during training. As a remedy, DST methods \citep{bellec2017deep, SET, mostafa2019parameter, SNFS, evci2020rigging, liu2021sparse, liu2021selfish, ITOP, graesser2022state} were introduced to train the neural networks under a given parameter budget while allowing mask change during training.

\subsection{Generative adversarial networks}
\textbf{Generative adversarial networks (GANs).} GANs \citep{goodfellow2020generative} have drawn considerable attention and have been widely investigated for years. 
% Various architectures have been proposed to enhance the capability of GANs.
Deep convolutional GANs \citep{radford2015unsupervised} replace fully-connected layers in the generator and the discriminator. Follow-up works \citep{gulrajani2017improved,karras2017progressive,biggan,zhang2019self} employed more advanced methods to improve the fidelity of generated samples. After that, several novel loss functions \citep{salimans2016improved,mao2017least,arjovsky2017wasserstein,  gulrajani2017improved, sun2020towards}, normalization and regularization methods \citep{SNGAN,terjek2019adversarial, wu2021gradient} were proposed to stabilize the adversarial training. Besides the efforts devoted to training GANs, image-to-image translation is also extensively explored \citep{zhu2017toward, zhu2017unpaired,choi2018stargan,karras2020analyzing,ledig2017photo,wang2018esrgan}. 
% Specifically, this direction includes semantic image synthesis \citep{zhu2017toward}, style transfer \citep{zhu2017unpaired,choi2018stargan,karras2020analyzing}, super resolution \citep{ledig2017photo,wang2018esrgan}, and such. 

\textbf{GAN balance.} Addressing the balance between the generator and discriminator in GAN training has been the focus of various works. However, directly applying existing methods to sparse GAN training poses challenges. For instance, \citep{arora2017generalization,bai2018approximability} offer theoretical analyses on the issue of imbalance but may have limited practical benefits, e.g., they require training multiple generators and discriminators. Empirically, BEGAN \citep{berthelot2017began} proposes to use proportional control theory to maintain a hyper-parameter $\frac{\mathbb{E}[|G(z)-D(G(z))|^\eta]}{\mathbb{E}[|x-D(x)|^\eta]}$, but it is only applicable when the discriminator is an auto-encoder. Unbalanced GAN \citep{ham2020unbalanced} pretrains a VAE to initialize the generator, which may only address the unbalance near initialization. GCC \citep{li2021revisiting} considers the balance during GAN compression, while its criterion requires a trained (dense) GAN, which is not given in the DST setting. Finally, STU-GAN \citep{liu2022don} proposes to use DST to address the unbalance issues but may fail under certain conditions, as demonstrated in our experiments.

\textbf{GAN compression and pruning.} 
% Like other deep neural networks, the training and inference process of GANs requires massive resource consumption and memory. 
One of the promising ways is based on neural architecture search and distillation algorithm \citep{li2020gan,fu2020autogan,hou2021slimmable}. Another part of the work applied pruning-based methods for generator compression \citep{shu2019co,yu2020self,jin2021teachers}. Later, works by \citep{wang2020gan} presented a unified framework by combing the methods mentioned above. Nevertheless, they only focus on the pruning of generators, thus potentially posing a negative influence on Nash Equilibrium between generators and discriminators. GCC \citep{li2021revisiting} compresses both components of GANs by letting the student GANs also learn the losses. Another line of work \citep{kalibhat2021winning, chen2021gans, chen2021data} tries to test the existence of lottery tickets in GANs. 
% However, most mentioned methods require fully-trained dense GANs in advance, hence being unable to reduce the training costs. 
To the best of our knowledge, STU-GAN \citep{liu2022don} is the only work that tries to directly train sparse GANs from scratch.

\section{Preliminary and setups}
Generative adversarial networks (GANs) have two fundamental components, a generator $G(\cdot;\bm{\theta}_{G})$ and a discriminator $D(\cdot;\bm{\theta}_{D})$. Specifically, the generator maps a sampled noise $\bm{z}$ from a multivariate normal distribution $p(\bm{z})$ into a fake image to cheat the discriminator. In contrast, the discriminator distinguishes the generator's output and the real images $\bm{x}_{r}$ from the distribution $p_\text{data}(\bm{x}_r)$. 

Formally, the optimization objective of the two-player game defined in GANs can be generally defined as follows:
\begin{align}
        \mathcal{L}_D (\bm{\theta}_{D}, \bm{\theta}_{G}) = &\mathbb{E}_{\bm{x}_{r} \sim p_\text{data}} \left [ f_1 (D(\bm{x}_{r};\bm{\theta}_{D})) \right ] + 
        \mathbb{E}_{\bm{z} \sim p} \left [ f_2 (D(G(\bm{z};\bm{\theta}_{G}))) \right ] \\
        \mathcal{L}_G (\bm{\theta}_{G}) = 
        &\mathbb{E}_{\bm{z} \sim p} \left [ g_1 (D(G(\bm{z};\bm{\theta}_{G}))) \right ].
\end{align}

To be more specific, different losses can be used, including the loss in the original JS-GAN \citep{goodfellow2020generative} where $f_1(x) = -\log(x)$, $f_2(x)=-g_1(x)=-\log(1-x)$; Wasserstein loss \citep{gulrajani2017improved} where $f_1(x) = -f_2(x) =g_1(x)= -x$; and hinge loss \citep{SNGAN} where $f_1(x)=\max(0,1-x)$, $f_2(x)=\max(0,1+x)$, and $g_1(x)=-x$. 
The {two components} are optimized alternately to achieve the Nash equilibrium. 

\textbf{GAN sparse training.} 
In this work, we are interested in sparse training for GANs. In particular, 
the objective of sparse GAN training can be formulated as follows: 
\begin{align}
        \bm{\theta}^*_{D}  &= \min_{\bm{\theta}_{D}} \mathcal{L}_D(\bm{\theta}_{D} \odot \bm{m}_D, \bm{\theta}_{G} \odot \bm{m}_G)   \nonumber \\ 
        \bm{\theta}^*_{G}  &= \min_{\bm{\theta}_{G}} \mathcal{L}_G(\bm{\theta}_{G} \odot \bm{m}_G)   \nonumber \\
\textrm{s.t.} \quad \bm{m}_D \in \{0,1\}^{\left|\bm{\theta}_D\right|}, 
        \quad \bm{m}_G &\in \{0,1\}^{\left|\bm{\theta}_G\right|}, \quad  
    \|\bm{m}_D\|_0 /{\left|\bm{\theta}_D\right|} \leq d_D, 
    \quad \|\bm{m}_G\|_0 /{\left|\bm{\theta}_G\right|} \leq d_G \nonumber, 
\end{align}

where $\odot$ is the Hadamard product; $\bm{\theta}^*_{D}$, $\bm{m}_{D}$, ${\left|\bm{\theta}_D\right|}$, $d_D$ are the sparse solution, mask, number of parameters, and target density for the discriminator, respectively. The corresponding variables for the generator are denoted with subscript $G$. For pruning-at-initialization methods, masks $\bm{m}$ are determined before training, whereas $\bm{m}$ are dynamically adjusted for dynamic sparse training (DST) methods.

\textbf{Dynamic sparse training (DST).} DST methods \citep{SET, evci2020rigging} usually start with a sparse network parameterized by $\bm{\theta} \odot \bm{m}$ with randomly initialized mask $\bm{m}$. After a constant time interval $\Delta T$, it updates mask $\bm{m}$ by removing a fraction of connections and activating new ones with a certain criterion. The total number of active parameters $\|\bm{m}\|_0$ is hence kept under a certain threshold $d |\bm{\theta}|$. Please see \autoref{sect:DST-hyperparams} for more details.

\section{Motivating observations: The unbalance in sparse GAN training}
\label{sect:gan_balance}
% In this section, we introduce the motivation and the formulation of the balance ratio.

% \subsection{The unbalance of GAN during training}

% TODO: Add related works
% It is non-trivial to maintain balance for sparse GAN training.
As discussed in \autoref{sect:intro}, it is essential to maintain the balance of generator and discriminator during GAN training. As strong discriminators may lead to over-fitting, whereas weak discriminators may be unable to detect mode
collapse. When it comes to sparse GAN training, the consequences caused by the unbalance can be further amplified. For example, sparsifying a weak generator while keeping the discriminator unmodified may lead to an even more unbalanced worst-case scenario.

\begin{figure*}[t]
  \centering
    \includegraphics[width=0.85\textwidth]{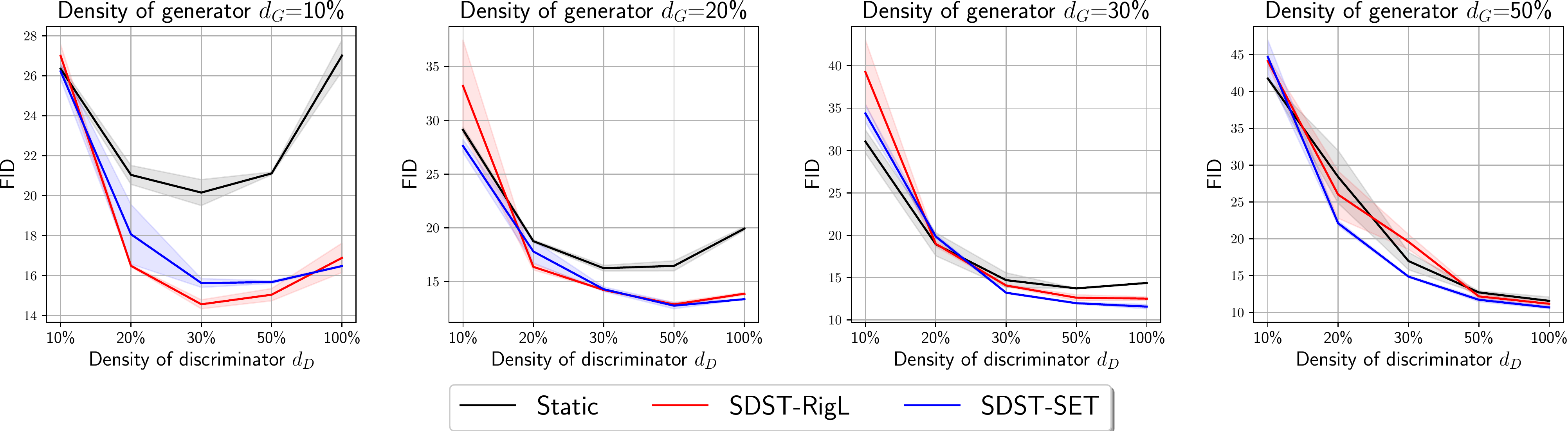}
   \caption{FID ($\downarrow$) comparison of \val{SDST} against \val{STATIC} sparse training for SNGAN on CIFAR-10 with different sparsity ratio combinations. The shaded areas denote the standard deviation.}
   \label{fig:baseline12-FID}
   \vspace{-0.5cm}
\end{figure*}

% Use experiments to support our claims
To support our claim, we conduct experiments with SNGAN \citep{SNGAN} on the CIFAR-10 dataset. We consider the following sparse training algorithms:
% \begin{itemize}

\ding{202} \textbf{Static sparse training (\val{STATIC})}. For \val{STATIC}, layer-wise sparsity ratio and masks $\bm{m}_G, \bm{m}_D$ are fixed throughout the training.
    
\ding{203} \textbf{Single dynamic sparse training (\val{SDST})}. \val{SDST} is a direct application of the DST method on GANs where only the generator dynamically adjusts masks during the training. We name such method \val{SDST} as only one component of the GAN, i.e., the generator, is dynamic. Furthermore, we call the variant which grows connections based on gradients magnitude as \rigl{SDST-RigL} \citep{evci2020rigging}, and randomly as \set{SDST-SET} \citep{SET}. Note that STU-GAN \citep{liu2022don} is almost identical to \rigl{SDST-RigL} with EMA \citep{EMA} tailored for DST. We do not consider naively applying DST on both generators and discriminators, as in STU-GAN, it is empirically shown that simply adjusting both components generates worse performance with more severe training instability.\footnote{We also perform a small experiment in \autoref{sect:appendix-double_sdst} to validate their findings.}
% \end{itemize}

% Following STU-GAN \citep{liu2022don}, 

We test the considered algorithms with $d_G \in \{10\%, 20\%, 30\%, 50\%\}$ and $d_D \in \{10\%, 20\%, 30\%, $ $ 50\%, 100\%\}$. More experiment details can be found at \autoref{sect:appendix-experiment-detail} and \autoref{sect:DST-hyperparams}. 
% are determined using \textit{Erd\H{o}s-R\'enyi-Kernel} (ERK) graph topology \citep{evci2020rigging} and 
% are fixed throughout the training. 

% \textbf{Experiment results.}
\subsection{Key observations}
We report the results in \autoref{fig:baseline12-FID} and summarize our critical findings as follows:

\ding{202} \textbf{Observation 1: Neither strong nor weak sparse discriminators can provide satisfactory results.} The phenomenon is most noticeable when $d_G=10\%$, where the FID initially decreases but then increases. The reasons may be as follows: (1) Overly weak discriminators may cause training collapse as they cannot provide useful information to the generator, resulting in a sudden increase in FID at the early stage of sparse GAN training. (2) Overly strong discriminators may not yield good FID results because they learn too quickly, not allowing the generator to keep up. Hence, to ensure a balanced training of GAN for sparse training methods, it is crucial to find an appropriate sparsity ratio for the discriminator.

\ding{203} \textbf{Observation 2: \val{SDST} is unable to give stable performance boost compared to the \val{STATIC} baseline.} Another critical observation is that \val{SDST} is better than \val{STATIC}  only when the discriminator is strong enough. More specifically, for all selected discriminator density ratios, \val{SDST} method is not better than \val{STATIC} when using a small discriminator density ($d_D = 10\%$). On the contrary, for the cases where $d_D > d_G$, we {generally} see a significant performance boost brought by \val{SDST}.

\section{Balance ratio: Towards quantifying the unbalance in sparse GAN training}
\label{sect:balance_ratio_def}

\subsection{Formulation of the balance ratio}

\begin{figure}[t]
  % \centering
  %   \includegraphics[width=0.9\textwidth]{plots/baseline1_BR.pdf}
  %  \caption{Balance ratio of \val{STATIC} baseline trained SNGAN on CIFAR-10 with different sparsity ratio combinations.}
  %  \label{fig:baseline1-BR}
  % \begin{figure*}[t]
  \centering
    \includegraphics[width=0.99\textwidth]{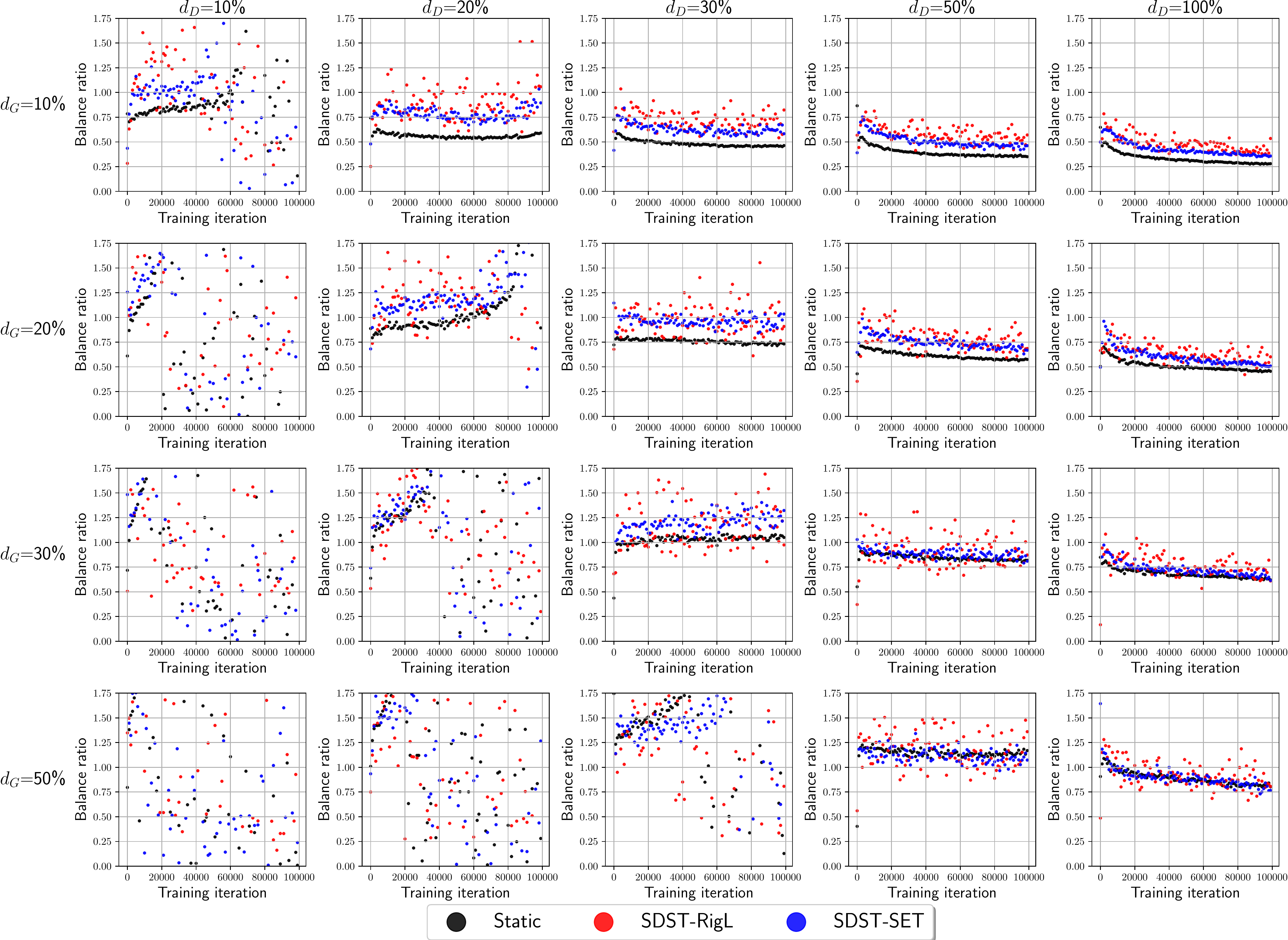}
   \caption{BR comparison of \val{SDST} against \val{STATIC} sparse training for SNGAN on CIFAR-10 with different sparsity ratio combinations. }
   \label{fig:baseline12-BR}
   % \vspace{-0.5cm}
% \end{figure*}
\end{figure}

To gain a deeper understanding of the phenomenon observed in the previous section, and to better monitor and control the degree of unbalance in sparse GANs, we introduce a novel quantity called the balance ratio (BR). This quantity is defined as follows.

At each training iteration, we draw random noise $\boldsymbol{z}$ from a multivariate normal distribution and real images $\boldsymbol{x}_r$ from the training set. We denote the discriminator after gradient descent update as $D(\cdot;\boldsymbol{\theta}_D)$. We denote generator before and after gradient descent training as $G^\text{pre}(\cdot; \boldsymbol{\theta}_G)$ and $G^\text{post}(\cdot; \boldsymbol{\theta}'_G)$, respectively. Then the balance ratio is defined as:
\begin{align}
    \label{eq: BR}
    \text{BR}=\frac{\mathbb{E}_{\bm{z} \sim p}\left[D(G^\text{post}(\boldsymbol{z}))-D(G^\text{pre}(\boldsymbol{z}))\right]}{\mathbb{E}_{\bm{x}_{r} \sim p_\text{data}}[D(\boldsymbol{x}_r)]-\mathbb{E}_{\bm{z} \sim p}[D(G^\text{pre}(\boldsymbol{z}))]} = \frac{\alpha}{\beta}.
\end{align}

\begin{wrapfigure}[11]{r}{0.5\linewidth}
\vspace{-5mm}
\begin{center}
\centerline{\includegraphics[width=0.5\columnwidth]{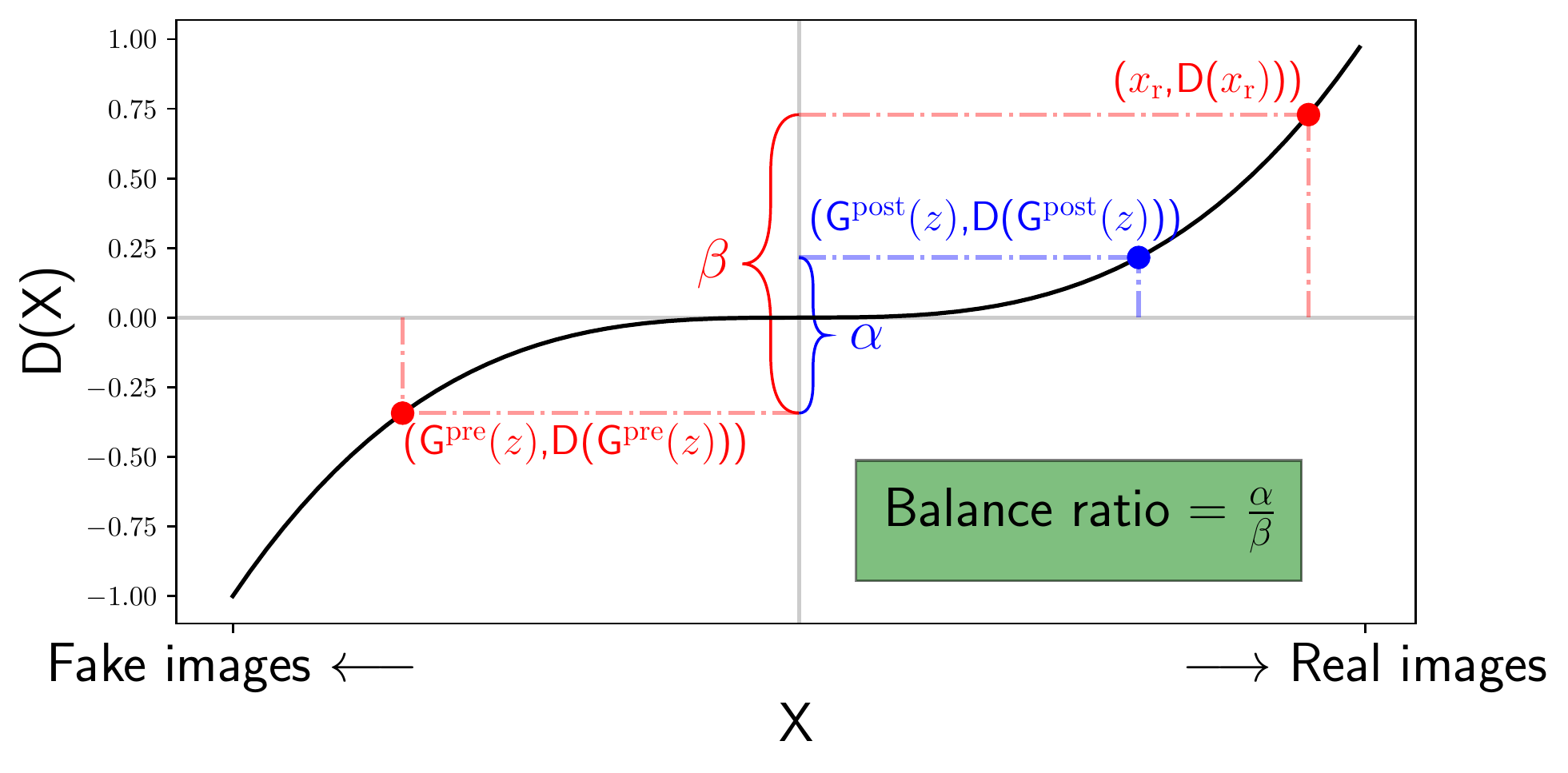}}
\caption{Illustration of balance ratio.}
\label{fig:illustration-BR}
\end{center}
\vskip -1cm
\end{wrapfigure}

Precisely, BR measures how much improvement the generator can achieve in the scale measured by the discriminator for a specific random noise $\boldsymbol{z}$. When BR is too small (e.g., BR$<30\%$), the updated generator is too weak to trick the discriminator, as the generated images are still considered fake. Similarly, for the case where BR is too large (e.g., BR$>80\%$), the discriminator is considered too weak hence it may not provide useful information to the generator. We also illustrate BR in \autoref{fig:illustration-BR}.

\subsection{Understanding observation 1: Analysing GAN balance with the balance ratio}
\label{sect:understand-ob1}

We visualize the BR evolution throughout the training for the experiments in \autoref{sect:gan_balance} to show the effectiveness of BR in quantifying the balance of sparse GANs. We show the results in \autoref{fig:baseline12-BR}.

It illustrates that BR can distinguish the density difference (hence the representation power difference) of the discriminator. Specifically, we can see that for larger discriminator density $d_D$, the BR is much lower throughout the training, indicating strong discriminators. On the contrary, for the cases where the discriminators are too weak compared to the generators, e.g., all cases where $d_D=10\%$, we can observe BR first increases and then oscillates wildly. We believe this oscillatory behavior is related to the training collapse. Empirical results also show that the FID metric experiences a sudden increase after this turning point.

% \subsection{Dynamic density adjustment of the discriminators}
\subsection{{Dynamic density adjust: A first attempt to utilize the balance ratio}}
\label{sect:DA}

As demonstrated in the previous section, the balance ratio (BR) effectively captures the degree of balance between the generators and discriminators in sparse GANs. Hence, it is natural to leverage BR to dynamically adjust the density of discriminators during sparse GAN training so that a reasonable discriminator density can be found. 

{To demonstrate the value of BR, we propose a simple yet powerful modification to the \val{STATIC} baseline. This method, which we call \textbf{dynamic density adjust (\val{DDA})}, is explained below.} Specifically, we initialize the initial density of the discriminator $d^\text{init}_D = d_G$. After a specific training iteration interval, we adjust the density of the discriminator based on the BR over the last few iterations with a pre-defined density increment $\Delta d$. With a pre-defined BR bounds $[B_{-}, B_{+}]$, we decrease $d_D$ by $\Delta d$ when BR is smaller than $B_{-}$, and vise versa.  We show the algorithm in \autoref{sect:appendix-algorithms}  \autoref{algo:DA}.

\textbf{Comparison to ADA \citep{ADA}}. In this paragraph, we compare ADA and \val{DDA}. (1) Notice that \val{DDA} algorithm is orthogonal to ADA in a sense that StyleGAN2-ADA adjusts the data augmentation probability while \val{DDA} adjusts the discriminator density. (2) Moreover, the criterion used in \val{DDA}, i.e. BR, is very different from the criterion proposed in StyleGAN2-ADA, i.e. $r_v=\frac{\mathbb{E}\left[D(x_\text{train})\right]-\mathbb{E}\left[D(x_\text{val})\right]}{\mathbb{E}\left[D(x_\text{train})\right]-\mathbb{E}\left[D(G(z))\right]}$ and $r_t = \mathbb{E}\left[\text{sign} (D(x_\text{train}))\right]$. In particular, $r_v$ requires a separate validation set, while $r_t$ only quantifies the overfitting of the discriminator to the training set. (3) Another note is that \val{DDA} is a flexible framework, where its criterion, i.e. BR, can be potentially replaced by $r_v$, $r_t$, and such.

\begin{figure}[t!]
    \centering
    \includegraphics[width=0.99\textwidth]{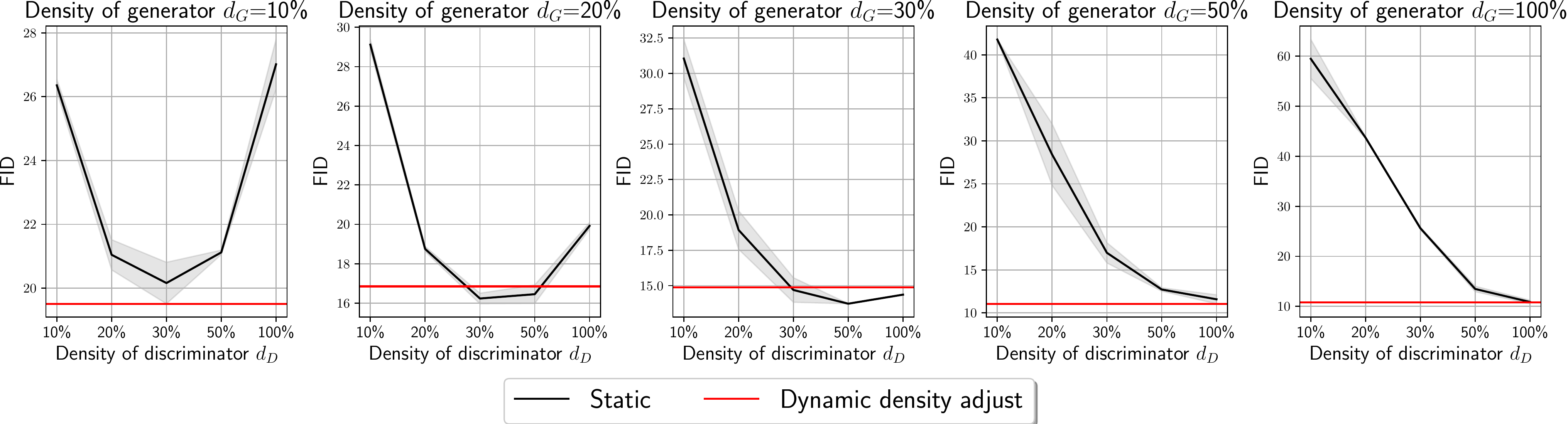}
   \caption{FID ($\downarrow$) of \val{STATIC} sparsely trained SNGAN with and without \val{DDA} on CIFAR-10 with different sparsity ratio combinations. The result of \val{DDA} is independent of $d_D$ as it is determined automatically. The shaded areas denote the standard deviation.}
   \label{fig:baseline1-FID}
\end{figure}

\textbf{Experiment results.} We test \val{DDA} with target BR interval $[B_{-}, B_{+}] = [0.5,0.65]$. Precisely, \val{DDA} tends to find a suitable discriminator where the generator can just trick the discriminator throughout the training. We show the results in \autoref{fig:baseline1-FID} with red lines. The experiments show that \val{DDA} can identify reasonable discriminator densities to enable balanced training for sparse GANs.

\subsection{Understanding observation 2: Analysing the failure of \val{SDST} with the balance ratio}
\label{sect:sdst}
% In this section, we investigate applying DST on GANs as STU-GAN \citep{liu2022don} implies that applying DST on the generator can partially address the training balance problem. 

% Like STU-GAN, we test a direct application of the DST method on GANs where only the generator dynamically adjusts masks during the training. We name such method \textbf{single dynamic sparse training (\val{SDST})} as only one component of the GAN, i.e., the generator, is dynamic. Furthermore, we call the variant which grows connections based on gradients magnitude as \rigl{SDST-RigL} \citep{evci2020rigging}, and randomly as \set{SDST-SET} \citep{SET}. Note that STU-GAN is almost identical to \rigl{SDST-RigL} with EMA \citep{EMA} tailored for DST. We do not consider naively applying DST on both generators and discriminators, as in STU-GAN, it is empirically shown that simply adjusting both components generates worse performance with more severe training instability.\footnote{We also perform a small experiment in \autoref{sect:appendix-double_sdst} to validate their findings.}

% We follow the same setting considered in \autoref{sect:gan_balance} where $d_G \in \{10\%, 20\%, 30\%, 50\%\}$\footnote{Note that DST makes no sense for the dense generator, i.e., $d_G=100\%$.} and $d_D \in \{10\%, 20\%, 30\%, 50\%, 100\%\}$. Detailed hyper-parameters can be found in \autoref{sect:DST-hyperparams}. We show the experiment results in \autoref{fig:baseline12-FID} along with \val{STATIC} baseline.

By leveraging BR, we can also gain further insights into why some configurations do not benefit from \val{SDST} as compared to \val{STATIC}.

\textbf{Regarding \val{SDST} as a way of increasing the generator capacity.} Our findings suggest that \val{SDST} can possibly enhance the generator's  representation power, as demonstrated by the higher BR values  compared to \val{STATIC} observed in \autoref{fig:baseline12-BR}. We attribute this effect to the in-time over-parameterization (ITOP) \citep{ITOP} induced by dynamic sparse training.

% By leveraging BR, we can also gain further insights into why some configurations do not benefit from \val{SDST} as compared to \val{STATIC}.

% Regarding \val{SDST} as a way of increasing the generator capacity. Our figure illustrates that \val{SDST} has higher BR compared to \val{STATIC}. We believe that the increase of the generator representation power is due to the in-time over-parameterization \citep{ITOP} induced by DST. 

\textbf{\val{SDST} does not address training collapse.} The increase in generator's representation power resulting from \val{SDST} is only beneficial when the discriminator has matching or superior representation power. Therefore, if the training has already collapsed for the static baseline methods (\val{STATIC}), meaning that the generator is already stronger than the discriminator, \val{SDST} may not be effective in stabilizing sparse GAN training. This is evident from the results shown in \autoref{fig:baseline12-BR} first-row column 1, second-row column 1, third-row columns 1-2, and fourth-row columns 1-3.

% Since \val{SDST} makes the generator stronger, such an increase is beneficial only when the discriminator has matching or better representation power. Hence, \val{SDST} is unable to stabilize sparse GAN training when the training collapse has already occurred for \val{STATIC} methods, i.e., the generator is already stronger than the discriminator. As is shown by first-row column 1; second-row column 1; third-row columns 1-2; fourth-row columns 1-3.

% {For the case where the generator is already stronger than the discriminator, using \val{SDST} is unable to address the training collapse problem.}

Despite the superior performance of STU-GAN (or \val{SDST} in general) at higher discriminator density ratios $d_D$, there exist some limitations for \val{SDST}, which we summarize below:

\ding{202} \val{SDST} requires a pre-defined discriminator density $d_D$ before training. However, it is unclear what is a good choice. In real-world scenarios, it is not practical to manually search for the optimal $d_D$ for each $d_G$. A workaround may be using the maximum allowed density for the discriminator. However, as shown in \autoref{fig:baseline12-FID}, the best performance is not always obtained with the maximum $d_D=100\%$. Moreover, we are wasting extra computational cost for a worse performance if we use an overly-strong discriminator.

\ding{203} \val{SDST} fails if there is an additional constraint on the density of the discriminator $d_D$. As \autoref{fig:baseline12-FID} suggests, for weak discriminators, \val{SDST} is unable to show consistent improvement compared to the \val{STATIC} baseline.

Hence, STU-GAN (or \val{SDST} in general), which directly applies DST to the generator, may only be useful when the corresponding discriminator is strong enough. In this sense, obtaining balanced training automatically is essential in GAN DST to deal with more complicated scenarios.

\section{Balanced dynamic sparse training for GANs}
In this section, we describe our methodology for balanced sparse GAN training.

STU-GAN (or \val{SDST} in general) considered in the last section cannot generate stable and satisfying performance. This implies that we should utilize the discriminator in a better way rather than do nothing (like \val{SDST}) or directly apply DST to the discriminator (see \autoref{sect:appendix-double_sdst} for additional experiments). Consequently, \val{DDA} (\autoref{sect:DA}), which adjusts the discriminator density to stabilize GAN training, is a favorable candidate to address the issue. To this end, we propose \textbf{b\underline{A}lanced \underline{D}yn\underline{A}mic s\underline{P}arse \underline{T}raining (\val{ADAPT})}, which adjusts the density of the discriminator during training with \val{DDA} while the generator performs DST. 

We further introduce two variants, namely \valsub{ADAPT}{relax} and \valsub{ADAPT}{strict}, based on whether we force the discriminator to be sparse. We present them in \autoref{sect:rddst} and \autoref{sect:sddst}. These methods are more flexible and generate more stable performance compared to \val{SDST}.
% \val{ADAPT}$_\text{relax}$:
\subsection{\val{ADAPT}$_\text{relax}$: Balanced dynamic sparse training in the relaxed setting}
\label{sect:rddst}

In this section, we consider the relaxed setting where a dense discriminator can be used, i.e., $d_D \leq d_\text{max} = 100\%$. This relaxed scenario gives the greatest flexibility to the discriminator. However, it does not necessarily enforce the sparsity of the discriminator (hence, no computational savings for the discriminator) because the density of the discriminator can be as high as $100\%$.

% We first investigate
{For the relaxed setting, we} use the direct combination of \val{SDST} with \val{DDA}. Precisely, the generator is adjusted using DST as mentioned in \autoref{sect:gan_balance} while the density of the discriminator is dynamically adjusted with \val{DDA} as mentioned in \autoref{sect:DA}. We call such a combination \textbf{relaxed balanced dynamic sparse training (\valsub{ADAPT}{relax})}.  Please see \autoref{sect:appendix-algorithms} \autoref{algo:R-DDST} for more details.
 
 \textbf{Comparison to STU-GAN (or \val{SDST} in general)}. Compared to STU-GAN (or \val{SDST} in general), {which pre-defines and fixes the discriminator density during training}, the difference is that for \val{ADAPT}$_\text{relax}$, the density of the discriminator is adjusted during the training process automatically through real-time monitoring of the balance ratio. Given the initial discriminator density $d_D^\text{int}=d_G$, \val{ADAPT}$_\text{relax}$ increases the discriminator density if a stronger discriminator is needed, and vice versa.
%\val{ADAPT}$_\text{strict}$:
\subsection{\val{ADAPT}$_\text{strict}$: Balanced dynamic sparse training in the strict setting}
\label{sect:sddst}
Different from \autoref{sect:rddst}, we now consider {a strict setting} where there is an additional sparsity constraint on the discriminator density in this section, i.e., $d_D \leq d_\text{max}<100\%$.

\valsub{ADAPT}{relax} introduced in the previous section does not necessarily enforce sparsity for the discriminator, which provides less memory/training resources saving for larger generator density ratios. Note that the discriminator does not take advantage of DST to explore the structure of the dense network. Hence, we further present \textbf{strict balanced dynamic sparse training (\valsub{ADAPT}{strict})} in this section. This method allows the discriminator to perform DST in a controlled manner, which can lead to a better exploration of the dense network structure while maintaining the balance between the generator and the discriminator. We explain how \valsub{ADAPT}{strict} differs from \valsub{ADAPT}{relax} below:

\ding{202} \textbf{Capacity increase of the discriminator.} The essential difference lies when the observed BR is higher than $B_+$, which means we need a stronger discriminator. In this case, if the discriminator density is lower than the constraint, i.e., $d_D < d_\text{max}$, \valsub{ADAPT}{strict} will perform just like \valsub{ADAPT}{relax} to increase the discriminator density. However, if the discriminator is already the maximum density, i.e., $d_D = d_\text{max}$, the discriminator will alternatively perform DST as a way of increasing the discriminator capacity (See \autoref{sect:sdst} for intuition).
 
\ding{203} \textbf{Capacity decrease of the discriminator.} Similar to \valsub{ADAPT}{relax}, when the observed BR is lower than $B_-$, we will decrease the discriminator density.

Hence, \valsub{ADAPT}{strict} makes the discriminator adaptive both in the density level (through density adjustment) and the parameter level (through DST). The algorithm of \valsub{ADAPT}{strict} is shown in \autoref{sect:appendix-algorithms} \autoref{algo:S-DDST}.
\begin{table}
    \caption{FID ($\downarrow$) of different sparse training methods \textbf{with no constraint on the density of the discriminator}. Best results are in \textbf{bold}; second-best results are \underline{underlined}.}
    % \begin{subtable}{0.49\textwidth}
    % \caption{
    % SNGAN on CIFAR-10 and STL-10.}
    \label{tab:benchmark-full}
    \resizebox{\columnwidth}{!}{
    \begin{tabular}{ l cccc cccc cccc cccc}
    \hline
    \toprule
    Dataset  &\multicolumn{4}{c}{\textbf{CIFAR-10 (SNGAN)}} &\multicolumn{4}{c}{\textbf{STL-10 (SNGAN)}} &\multicolumn{4}{c}{\textbf{CIFAR-10 (BigGAN)}} &\multicolumn{4}{c}{\textbf{TinyImageNet (BigGAN)}}\\
    \midrule
    Generator density & 10\% & 20\%  & 30\% & 50\% & 10\% & 20\%  & 30\% & 50\%  & 10\% & 20\%  & 30\% & 50\%& 10\% & 20\%  & 30\% & 50\%\\
    \midrule
    (Dense Baseline)   & \multicolumn{4}{c}{10.74} & \multicolumn{4}{c}{29.71}  & \multicolumn{4}{c}{8.11}  & \multicolumn{4}{c}{15.43 } \\
    \midrule
    STATIC-Balance  & 26.75 & 19.04 & 15.05 & 12.24 & 48.18 & 44.67 & 41.73 & 37.68 & 16.98 & 12.81 & 10.33 & 8.47 & 28.78 & 21.67 & 18.86 & 17.51\\
    STATIC-Strong  & 26.79& 19.65 & 14.38 & 11.91 & 52.48 & 43.85 & 42.06 & 37.47& 23.48 & 14.26 & 11.19 & 8.64 & 31.44 & 22.51 & 18.22 & 18.00 \\
    \midrule
    \midrule
    \tikzdiamond[blue, fill=blue] SDST-Balance-SET  & 26.23 & 17.79 & 13.21 & 11.79 & 56.41 & 46.58 & 39.93 & 30.37 & 12.41 & 9.87 & 9.13 & \textbf{8.01}&25.39&21.30&21.80&21.20\\
    \tikzdiamond[blue, fill=blue] SDST-Strong-SET  & \underline{16.49} & \underline{13.36} & \textbf{11.68} & \underline{10.68} & 67.37 & 49.96 & 37.99 & 31.08& 18.94 & 9.64 & 8.75 & 8.36 &22.20&20.56&21.70&18.32\\

    \tikzcircle[red,fill=red] SDST-Balance-RigL  &  27.06 & 16.36 & 14.00 & 12.28 & \underline{43.08} & 33.90 & 31.83 & 30.30 & 12.45 & 9.42 & 8.86 & \underline{8.03} & 21.60 & 19.33 & 18.57 & 17.45\\
    \tikzcircle[red,fill=red] SDST-Strong-RigL  & 17.02 & 13.86 & 12.51 & 11.35 & 53.65 & \underline{33.25} & \textbf{31.41} & \underline{30.18}& \underline{10.58} & \underline{9.11} & \underline{8.69} & 8.33 & \underline{21.14} & \underline{18.95} & \underline{17.75} & \underline{16.30}\\

    \midrule
    \midrule
    $\text{ADAPT}_\text{relax}$ (Ours)  & \textbf{14.19} & \textbf{13.19} & \underline{12.38} & \textbf{10.60} & \textbf{35.98} & \textbf{33.06} & \underline{31.71} & \textbf{29.96}& \textbf{10.19} & \textbf{8.56} & \textbf{8.36} & 8.22 & \textbf{19.42} & \textbf{17.99} & \textbf{17.06} & \textbf{14.15}\\
    
    \bottomrule

    \end{tabular}
    }
    \vspace{-0.5cm}
\end{table}

\begin{table}
    \centering
    \caption{
    FID ($\downarrow$) of different sparse training methods. \textbf{The density of the discriminator is constrained to be lower than 50\%}. Best results are in \textbf{bold}; second-best results are \underline{underlined}.}
    \label{tab:benchmark-sparse}
    \resizebox{0.99 \columnwidth}{!}{
    \begin{tabular}{ l cccc cccc cccc cccc}
    \\
    \hline
    \toprule
    Dataset  &\multicolumn{4}{c}{\textbf{CIFAR-10 (SNGAN)}} &\multicolumn{4}{c}{\textbf{STL-10 (SNGAN)}} &\multicolumn{4}{c}{\textbf{CIFAR-10 (BigGAN)}} &\multicolumn{4}{c}{\textbf{TinyImageNet (BigGAN)}}\\
    \midrule
    Generator density & 10\% & 20\%  & 30\% & 50\% & 10\% & 20\%  & 30\% & 50\% & 10\% & 20\%  & 30\% & 50\% & 10\% & 20\%  & 30\% & 50\% \\
    % \midrule
    % Sparsity percentage & 89.26\% & 95.60\% &  89.26\% & 95.60\% \\ 
    \midrule
    (Dense Baseline)   & \multicolumn{4}{c}{10.74} & \multicolumn{4}{c}{29.71}  & \multicolumn{4}{c}{8.11}  & \multicolumn{4}{c}{15.43 } \\
    \midrule
    STATIC-Balance  & 26.75 & 19.04 & 15.05 & 12.58 & 48.18 & 44.67 & 41.73 & 37.68 & 16.98 & 12.81 & 10.33 & 8.47 & 28.78 & 21.67 & 18.86 & 17.51\\
    STATIC-Strong  & 21.73 & 16.69 & 13.48 & 12.58 & 50.36 & 44.06 & 40.73 & 37.68 & 18.91 & 13.43 & 10.84 & 8.47 & 33.01 & 23.93 & \underline{17.90} & 17.51\\

    \midrule
    \midrule
    \tikzdiamond[blue, fill=blue] SDST-Balance-SET  &  26.23 & 17.79 & 13.21 & \textbf{11.79} & 56.24 & 44.51 & 41.23 & 30.80 & 12.41 & 9.87 & 9.13 & \underline{8.01} &25.39&21.30&21.80&21.20\\
    \tikzdiamond[blue, fill=blue] SDST-Strong-SET  & 15.68 & \underline{12.75}&\textbf{11.98}&\textbf{11.79} &57.91&50.05&38.13&30.80 &11.85 & 9.39 & 8.61 & \underline{8.01} &22.68&20.24&22.00&21.20\\
    \tikzcircle[red,fill=red] SDST-Balance-RigL  &  27.06 & 16.36 & 14.00 & 12.28 & \underline{43.08} & \underline{33.90} & \underline{31.64} & \underline{30.30} & 12.45 & 9.42 & 8.86 & 8.03 & \underline{21.60} & \underline{19.33} & 18.57 & \underline{17.45}\\
    \tikzcircle[red,fill=red] SDST-Strong-RigL  &  \underline{15.19} & 12.93 & 12.75 & 12.28 & 53.74 & 37.34 & 31.98 & \underline{30.30} &\underline{10.11} & \underline{9.17} & \textbf{8.35} & 8.03 & 21.90 & 20.43 & 18.29 & \underline{17.45}\\
    \midrule
    \midrule
    % \tikzdiamond[blue, fill=blue] $\text{ADAPT}_\text{strict}$ (SET)  &\\
    $\text{ADAPT}_\text{strict}$ (Ours)  & \textbf{14.53} & \textbf{12.73} & \underline{12.20} & \underline{12.11} & \textbf{41.18} & \textbf{31.59} & \textbf{31.16} & \textbf{29.11} & \textbf{9.29} & \textbf{8.64} & \underline{8.44} & \textbf{7.90}  & \textbf{18.89} & \textbf{17.37} & \textbf{16.93} & \textbf{16.02}\\

    % \midrule
    % \midrule
    % NTK-SAP (Ours) &\textbf{58.87}&\textbf{49.43}&\textbf{68.28}&\textbf{60.79} \\
    \bottomrule

    \end{tabular}
    }
    \vspace{-0.5cm}
\end{table}

\subsection{Experiment setting}
\textbf{Datasets, architectures, and target sparsity ratios.} We conduct experiments on SNGAN with ResNet architectures on the CIFAR-10 and the STL-10 \citep{stl10} datasets. We have also conducted experiments with BigGAN \citep{biggan} on the CIFAR-10 and TinyImageNet dataset (with DiffAug \citep{zhao2020differentiable}). Target density ratios of the generators $d_G$ are chosen from $\{10\%, 20\%, 30\%, 50\%\}$. Please see \autoref{sect:appendix-experiment-detail} for more experiment details.

\textbf{Baseline methods and two practical strategies.} We use \val{STATIC}  and \val{SDST} (\autoref{sect:gan_balance}) as our baselines. Note that in real-world application scenarios, it is not practical to perform a grid search for a good $d_D$ as in \autoref{sect:gan_balance}. Hence, we propose two practical strategies to define the constant discriminator density for these baseline methods: (1) balance strategy, where we set the density of the discriminator $d_D$ the same as the density of the generator $d_G$; (2) strong strategy, where we set the density of the discriminator as large as possible, i.e., $d_D=d_\text{max}$. For \val{SDST} methods, we test both \tikzdiamond[blue, fill=blue] \val{SDST-SET} and \tikzcircle[red,fill=red] \val{SDST-RigL}. For a fair comparison, $d_\text{max}$ is set to be 100\% and 50\% for the relaxed setting and the strict setting, respectively.

% baseline methods can use the discriminator with arbitrary sparsity ratio, i.e., $d_D \in [d_\text{min}, d_\text{max}]=[0\%, 100\%]$.

For \valsub{ADAPT}, we use the \rigl{RigL} version, which grows connections of the generators and discriminators based on gradient magnitude. The gradient information enables two components to react promptly according to the change of each other. Different from the value used in \autoref{sect:DA}, we control the balance ratio in the range $\left[0.45, 0.55\right]$ {unless otherwise mentioned} to have a slightly stronger discriminator, potentially avoiding training collapse. More details can be found in \autoref{sect:DST-hyperparams}. 

\subsection{Experiment results}
\begin{table}
    \caption{Normalized training FLOPs ($\downarrow$) of different sparse training methods with no constraint on the density of the discriminator. }
    \label{tab:benchmark-full-flops}
    \resizebox{\columnwidth}{!}{
    \begin{tabular}{ l cccc cccc cccc cccc}
    \hline
    \toprule
    Dataset  &\multicolumn{4}{c}{\textbf{CIFAR-10 (SNGAN)}} &\multicolumn{4}{c}{\textbf{STL-10 (SNGAN)}} &\multicolumn{4}{c}{\textbf{CIFAR-10 (BigGAN)}} &\multicolumn{4}{c}{\textbf{TinyImageNet (BigGAN)}}\\
    \midrule
    Generator density & 10\% & 20\%  & 30\% & 50\% & 10\% & 20\%  & 30\% & 50\%  & 10\% & 20\%  & 30\% & 50\%& 10\% & 20\%  & 30\% & 50\%\\
    \midrule
    (Dense Baseline)   & \multicolumn{4}{c}{100\% ($2.67 \times 10^{17}$)} & \multicolumn{4}{c}{100\% ($3.94 \times 10^{17}$)}  & \multicolumn{4}{c}{100\% ($6.81 \times 10^{17}$)}  & \multicolumn{4}{c}{100\% ($9.85 \times 10^{17}$)} \\
    \midrule
    Static-Balance   
    & 8.97\% & 17.08\% & 26.25\% & 47.25\%
    & 27.30\% & 47.14\% & 59.22\% & 73.35\%
    & 9.79\% & 19.02\% & 28.66\% & 49.03\%
    & 23.25\% & 44.87\% & 60.91\% & 79.29\%
    \\
    Static-Strong  
    & 58.29\% & 60.94\% & 64.53\% & 74.61\%
    & 86.12\% & 86.94\% & 87.60\% & 88.84\%
    & 83.78\% & 84.80\% & 86.21\% & 90.15\%
    &  48.02\% & 61.62\% & 72.79\% & 85.79\%
    \\
    \midrule
    \midrule
    \tikzdiamond[blue, fill=blue] SDST-Balance-SET  
    & 9.78\% & 18.91\% & 28.35\% & 48.44\%
    & 27.55\% & 47.60\% & 60.17\% & 75.38\%
    & 10.35\% & 20.12\% & 29.96\% & 49.82\%
    & 21.13\%& 37.06\%&48.83\%&65.58\%
    \\

    \tikzdiamond[blue, fill=blue] SDST-Strong-SET  
    &59.25\% & 62.94\% & 66.89\% & 75.96\%
    & 86.36\% & 87.43\% & 88.49\% & 90.82\%
    & 84.36\% & 85.90\% & 87.52\% & 90.95\%
    & 45.66\% & 53.91\% & 60.61\% & 71.88\%
    \\

    \tikzcircle[red,fill=red] SDST-Balance-RigL  
    & 10.71\% & 17.43\% & 25.66\% & 43.56\%
    & 29.51\% & 50.41\% & 63.34\% & 79.03\%
    & 9.92\% & 19.30\% & 28.90\% & 48.31\%
    & 24.97\% & 43.86\% & 57.26\% & 76.75\%
    
    \\
    \tikzcircle[red,fill=red] SDST-Strong-RigL  
    & 58.63\% & 61.35\% & 64.04\% & 71.01\%
    & 88.51\% & 90.24\% & 91.78\% & 94.57\%
    & 83.97\% & 85.24\% & 86.59\% & 89.54\%
    & 50.05\% & 61.02\% & 69.64\% & 83.35\%
    \\

    \midrule
    \midrule
    $\text{ADAPT}_\text{relax}$ (Ours) 
    & 36.67\% & 57.62\% & 61.31\% & 70.11\%
    & 46.73\% & 77.92\% & 83.62\% & 90.49\%
    & 10.39\% & 25.90\% & 40.65\% & 80.76\%
    & 29.75\% & 51.98\% & 64.57\% & 80.81\%
    \\
    
    \bottomrule

    \end{tabular}
    }
    \vspace{-0.5cm}
\end{table}

We show the experiment results in \autoref{tab:benchmark-full} and \autoref{tab:benchmark-sparse} for the relaxed setting and the strict setting, respectively. We also present the training FLOPs normalized by the dense counterpart for the relaxed setting in \autoref{tab:benchmark-full-flops}. We defer the results for the strict setting to \autoref{sect:appendix-flops-accurate} \autoref{tab:benchmark-sparse-flops}. We show FID for the CIFAR-10 test set, Inception scores, and comparison with post-hoc pruning baseline in \autoref{sect:appendix-more-results}. We also show \val{ADAPT} BR evolution  in \autoref{sect:DDST-BR}.  We summarize our findings below.

% More detailed training costs comparison can be found in \autoref{sect:appendix-flops-accurate}.

% \subsubsection{Experiment results}

\textbf{The strong strategy and the balance strategy for baselines.} Generally, using the strong strategy has some advantages over the balance strategy. Such an observation is most prominent in the CIFAR-10 dataset. For the cases where the balance strategy is better, e.g., SNGAN on the STL-10 dataset, our explanation is that the size difference between generators and discriminators is more significant. Hence, the degree of unbalance is more severe and leads to more detrimental effects.

\textbf{Comparison of \rigl{RigL} and \set{SET} for \val{SDST}.} We found that \rigl{RigL} has an advantage over \set{SET} when dealing with more sparse generators. Our hypothesis is that gradient information can effectively guide the generator to identify the most crucial connections in such cases. However, this advantage is not as apparent for more dense generators.

\textbf{\val{ADAPT}$_\text{relax}$ achieves a good trade-off between performance and computational cost.} Experiments show that \valsub{ADAPT}{relax} shows promising performance by being best for 13 out of 16 cases. The advantage of \valsub{ADAPT}{relax} is most prominent for the most difficult case, i.e., $d_G=10\%$. Specifically, it shows 2.3 and 7.1 FID improvements over the second-best methods for the SNGAN on the CIFAR-10 and the STL-10, respectively. Moreover, compared to the competitive baseline methods that use the strong strategy, i.e., \rigl{SDST-Strong-RigL} and \set{SDST-Strong-SET}, \valsub{ADAPT}{relax} shows great computational cost reduction. For example, it outperforms \rigl{SDST-Strong-RigL} on BigGAN (CIFAR-10) with much-reduced training FLOPs (10.39\% v.s. 83.97\%).

% \subsubsection{Main takeaway}
% This section compares \valsub{ADAPT}{relax} with other sparse training baselines. We find that \ding{202} \rigl{RigL} and the strong strategy are preferred compared to \set{SET} and the balance strategy for baseline methods. \ding{203} \rigl{SDST-Strong-RigL} ranks top consistently among the sparse training baselines. \ding{204} \valsub{ADAPT}{relax} shows promising performance by ranking top two for 11 out of 12 cases. More importantly, it shows better or comparable performance compared to \rigl{SDST-Strong-RigL} with much less computational cost.

% \subsection{Strict balanced dynamic sparse training}
% \label{sect:sddst}

% \textbf{Experiment setup.} We use the same baselines and adopt the same general setup in \autoref{sect:rddst} with the exception that $d_\text{max} = 50 \%$. We again use the \rigl{RigL} version for the \valsub{ADAPT}{strict}, which controls the balance ratio between $\left[0.45, 0.55\right]$. Please see \autoref{sect:DST-hyperparams} for more details.

% We present the results and normalized training FLOPs in \autoref{tab:benchmark-sparse}. We show IS and other results in \autoref{sect:appendix-more-results}. The BR evolution can be found in \autoref{sect:DDST-BR}.

% \subsubsection{Experiment results}

\textbf{\valsub{ADAPT}{strict} shows stable and superior performance.}
Similar to \valsub{ADAPT}{relax}, we notice that \valsub{ADAPT}{strict} also delivers promising results compared to baselines, even with a further constraint on the discriminator. More precisely, among all the cases, \valsub{ADAPT}{strict} ranks top 2 for all cases, with 13 cases being the best. Moreover, \valsub{ADAPT}{strict} again shows comparable or better performance compared to \rigl{SDST-Strong-RigL} with reduced computational cost.  

A more interesting observation is that \valsub{ADAPT}{strict} sometimes outperforms \valsub{ADAPT}{relax}. We speculate that this phenomenon occurs because changes in density may result in a larger influence on the GAN balance during training compared to DST. Hence, the strict version, whose discriminator density range is smaller,  may offer a more consistent performance.

% \subsubsection{Main takeaway}
 % This section reports the results for \valsub{ADAPT}{strict} with its baselines. We find that: \ding{202} Even when extra sparsity is introduced, \rigl{RigL} still demonstrates better results compared with \set{SET} for baseline methods in most experiments. \ding{203} For baseline methods, while the strong strategy shows favorable performance in the CIFAR-10 dataset, the gain is not salient for the STL-10 dataset. \ding{204} Most importantly, \valsub{ADAPT}{strict} again shows promising results by outperforming \rigl{SDST-Strong-RigL} in 11 out of 12 cases with much reduced computational costs. \ding{205} Lastly, \valsub{ADAPT}{strict} can sometimes have comparable or even better performance in some cases when compared to \valsub{ADAPT}{relax}.

% \input{table/table_biggan_strict.tex}
% \input{table/table_bigtiny_strict}
\section{Conclusion}
% In this paper, we study DST for GANs. We find that simply applying DST methods to the generator is not sufficient to improve the performance of sparse GANs. Hence, we propose to use BR to study the degree of unbalance between the sparse generators and the discriminators. We find that the application of DST on the generator is only beneficial when the discriminator is relatively stronger. Most importantly, we propose \val{ADAPT} to dynamically adjust the discriminator in both parameter and density levels. Our method demonstrates encouraging results. We believe our study may help researchers better understand the balance of two components in GAN training and encourage more researchers to investigate sparse training for GANs. It is also important to note that we have not yet evaluated our methods on the latest GAN architectures due to computational limitations.
In this paper, we investigate the use of DST for GANs and find that solely applying DST to the generator does not necessarily enhance the performance of sparse GANs. To address this, we introduce BR to examine the degree of unbalance between the sparse generators and discriminators. We find that applying DST to the generator only benefits the training when the discriminator is comparatively stronger. Additionally, we propose \val{ADAPT}, which can dynamically adjust the discriminator at both the parameter and density levels. Our approach shows promising results, and we hope it can aid researchers in better comprehending the interplay between the two components of GAN training and motivate further exploration of sparse training for GANs. However, we must note that we have not yet evaluated our methods on the latest GAN architectures due to computational constraints.

\section{Acknowledgement}
This work utilizes resources supported by the National Science Foundation’s Major Research Instrumentation program, grant No.1725729 \citep{kindratenko2020hal}, as well as the University of Illinois at Urbana-Champaign. This work is supported in part by Hetao Shenzhen-Hong Kong Science and Technology Innovation Cooperation Zone Project (No.HZQSWS-KCCYB-2022046); University Development Fund UDF01001491 from the Chinese University of Hong Kong, Shenzhen; Guangdong Key Lab on the Mathematical Foundation of Artificial Intelligence, Department of Science and Technology of Guangdong Province. Prof. NH is in part supported by Air Force Office of Scientific Research (AFOSR) grant FA9550-21-1-0411. 

\newpage

% \section*{References}

\bibliography{main}
\bibliographystyle{plainnat}

\newpage
\appendix

\section*{Overview of the Appendix}
The Appendix is organized as follows:
\begin{itemize}
    \item \autoref{sect:appendix-experiment-detail} introduces the general experimental setup.
    \item \autoref{sect:DST-hyperparams} introduces the details of dynamic sparse training.
    \item \autoref{sect:appendix-algorithms} shows detailed algorithms, i.e., \val{DDA}, \valsub{ADAPT}{relax}, and \valsub{ADAPT}{strict}.
    % \item \autoref{sect:SDST-STATIC-BR} gives detailed comparison of BR for \val{STATIC} and \val{SDST} in \autoref{sect:sdst}.
    \item \autoref{sect:DDST-BR} shows the BR evolution during training for \val{ADAPT}.
    \item \autoref{sect:appendix-more-results} shows additional results, including IS and FID of test sets of the main paper.
    \item \autoref{sect:appendix-flops-accurate} shows detailed FLOPs comparisons of sparse training methods.
\end{itemize}

\section{Experimental setup}
\label{sect:appendix-experiment-detail}
In this section, we explain the training details used in our experiments. Our code is mainly based on the original code of ITOP \citep{ITOP} and GAN ticket \citep{chen2021gans}. 

\subsection{Architecture details}
We use ResNet-32 \citep{resnet} for the CIFAR-10 dataset and ResNet-48 for the STL-10 dataset. See \autoref{tab:arch-cifar10} and \autoref{tab:arch-stl10} for detailed architectures. We apply spectral normalization for all fully-connected layers and convolutional layers of the discriminators.

For BigGAN architecture, we use the implementation used in DiffAugment \citep{zhao2020differentiable}.\footnote{\url{https://github.com/mit-han-lab/data-efficient-gans/tree/master/DiffAugment-biggan-cifar}.}

\subsection{Datasets}
We use the training set of CIFAR-10, the unlabeled partition of STL-10, and the training set of TinyImageNet for GAN training. Training images are resized to $32\times32$, $48 \times48$, $64 \times64$ for CIFAR-10, STL-10, and TinyImageNet datasets, respectively. Augmentation methods for both datasets are random horizontal flip and per-channel normalization. 

\subsection{Training hyperparameters}

\textbf{SNGAN on the CIFAR-10 and STL-10 datasets.} We use a learning rate of $2 \times 10^{-4}$ for both generators and discriminators. The discriminator is updated five times for every generator update. We adopt Adam optimizer with $\beta_1=0$ and $\beta_2=0.9$. The batch size of the discriminator and the generator is set to 64 and 128, respectively. Hinge loss is used following \citep{biggan, chen2021gans}. We use exponential moving average (EMA) \citep{EMA} with $\beta =0.999$. The generator is trained for a total of 100k iterations.

\textbf{BigGAN on the CIFAR-10 dataset.} We use a learning rate of $2 \times 10^{-4}$ for both generators and discriminators. The discriminator is updated four times for every generator update. We adopt Adam optimizer with $\beta_1=0$ and $\beta_2=0.999$. The batch size of both the discriminator and the generator is set to 50. Hinge loss is used following \citep{biggan, wu2021gradient}.  We use EMA with $\beta =0.9999$. The generator is trained for a total of 200k iterations.

{\textbf{BigGAN on the TinyImageNet dataset.} We use DiffAug \citep{zhao2020differentiable} to augment the input. The learning rate of the discriminator and the generator are set to $4 \times 10^{-4}$ and $1 \times 10^{-4}$, respectively. The discriminator is updated one time for every generator update. We adopt Adam optimizer with $\beta_1=0$ and $\beta_2=0.999$. The batch size of both the discriminator and the generator is set to 256. Hinge loss is used following \citep{biggan, wu2021gradient}.  We use EMA with $\beta =0.9999$. The generator is trained for a total of 200k iterations.}

\subsection{Evaluation metric}
\textbf{SNGAN on the CIFAR-10 and the STL-10 datasets.} We compute Fréchet inception distance (FID) and Inception score (IS) for 50k generated images every 5000 iterations. Best FID and IS are reported. For the CIFAR-10 dataset, we report both FID for the training set and test set, whereas, for the STL-10 dataset, we report the FID of the unlabeled partition.

\textbf{BigGAN on the CIFAR-10 and the TinyImageNet dataset.} We compute Fréchet inception distance (FID) and Inception score (IS) for 10k generated images every 5000 iterations. Best FID and IS are reported. 

\section{Dynamic sparse training details}
\label{sect:DST-hyperparams}
\subsection{How the generator performs DST}

In this section, we explain how the generator performs DST below. Note that the generator performs the same for \val{SDST} and \val{ADAPT}.

\textbf{Sparsity distribution at initialization.} Following RigL and ITOP \citep{evci2020rigging, ITOP}, only parameters of fully connected and convolutional layers will be pruned. At initialization, we use the commonly adopted \textit{Erd\H{o}s-R\'enyi-Kernel} (ERK) strategy \citep{evci2020rigging, SNFS, ITOP} to allocate higher sparsity to larger layers. Specifically, the sparsity of convolutional layers $l$ is scaled with $1-\frac{n^{l-1}+n^{l}+w^l+h^l}{n^{l-1}n^{l}w^{l}h^{l}}$, where $n^l$ denotes the number of channels of layer $l$ while $w^l$ and $h^l$ are the widths and the height of the corresponding kernel in that layer. For fully connected layers, \textit{Erd\H{o}s-R\'enyi} (ER) strategy is used, where the sparsity is scaled with $1-\frac{n^{l-1}+n^{l}}{n^{l-1}n^{l}}$.

\textbf{Update schedule.} The update schedule controls how many connections are adjusted per DST operation. It can be specified by the number of training iterations between sparse connectivity updates $\Delta T_G$, the initial fraction of connections adjusted ${\gamma}$, and decaying schedule $f_\text{decay}({\gamma}, T)$ for ${\gamma}$.  

\textbf{Drop and grow.} After $\Delta T_G$ training iterations, we update the mask $\bm{m}_G$ by dropping/pruning $f_\text{decay}({\gamma}, T) \left|\bm{\theta}_G\right| d_G$ number of connections with the lowest magnitude, where $\left|\bm{\theta}_G\right|$, $d_G$ are the number of parameters and target density for the generator, $f_\text{decay}({\gamma}, T)$ is the update schedule. Right after the connection drop, we regrow the same amount of connections. 

For the growing criterion, we test both random growth \tikzdiamond[blue, fill=blue] \val{SET} \citep{SET, ITOP} and gradient-based growth \tikzcircle[red,fill=red] \val{RigL} \citep{evci2020rigging}. Concretely, gradient-based methods find newly-activated connections $\theta$ with the highest gradient magnitude $\left|\frac{\partial \mathcal{L}}{\partial \theta}\right|$, while random-based methods explore connections in a random fashion. All the newly-activated connections are set to 0. One thing that should be noticed is that while previous works consider layer-wise connections drop and growth, we grow and drop connections globally as it grants more flexibility to the DST method.

\textbf{EMA for sparse GAN.} EMA \citep{EMA} is well-known for its ability to alleviate the non-convergence of GAN. We also implement EMA for sparse GAN training. Specifically, we zero out the moving average of dropped weights whenever there is a mask change. 

\subsection{DST hyperparameters for the generator}
\label{sect:DST-hyperparams-SDST}

We use the same hyper-parameters for \val{SDST} and \val{ADAPT}. The initial $\gamma$ is set to 0.5, and we use a cosine annealing function $f_\text{decay}$ following RigL and ITOP.

\textbf{SNGAN on the CIFAR-10 and the STL-10 datasets.} The connection update frequency of the generator $\Delta T_G$ is set to 500 and 1000 for the CIFAR-10 dataset and STL-10 dataset, respectively.  

\textbf{BigGAN on the CIFAR-10 and the TinyImageNet dataset.} The connection update frequency of the generator  $\Delta T_G$ is set to be 1000.

\subsection{Density dynamic adjust (\val{DDA}) hyper-parameters}
In this section, we provide hyper-parameters used in \autoref{sect:DA}. We set $d_D=2000$ , $\Delta T_D=0.05$, $\left[B_-, B_+\right] = [0.5, 0.65]$. Time-averaged BR over 1000 iterations is used as the indicator.

\subsection{DST hyperparameters for the discriminator in \val{ADAPT}}

We use a constant BR interval $[B_{-}, B_{+}] = [0.45, 0.55]$ for SNGAN experiments and BigGAN on the CIFAR-10 dataset. We set the BR interval $[B_{-}, B_{+}] = [0.3, 0.4]$ for BigGAN on the TinyImageNet since it uses DiffAug. Time-averaged BR over 1000 iterations is used as the indicator. Density increment $\Delta d$ is set to be 0.05, 0.025, and 0.05 for SNGAN (CIFAR-10), SNGAN (STL-10), and BigGAN (CIFAR-10), respectively. We use the same setting used in \autoref{sect:DST-hyperparams-SDST} for the generator. 

\textbf{Hyper-parameters for \valsub{ADAPT}{relax}.}  The density update frequency of the discriminator $\Delta T_D$ is 1000, 2000, 5000, and 10000 iterations for SNGAN (CIFAR-10), SNGAN (STL-10), BigGAN (CIFAR-10), and BigGAN (TinyImageNet), respectively. 

\textbf{Hyper-parameters for \valsub{ADAPT}{strict}.} The density/connections update frequency of the discriminator $\Delta T_D$ is 2000, 2000, 5000, and 10000 iterations for SNGAN (CIFAR-10), SNGAN (STL-10), BigGAN (CIFAR-10), and BigGAN (TinyImageNet), respectively.  

Note that we compute BR for every iteration to visualize the BR evolution, whereas one should note that such computational cost can be greatly decreased if BR is computed every few iterations.

\begin{table}
\begin{minipage}{.5\linewidth}
\caption{
ResNet architecture for CIFAR-10.}
\label{tab:arch-cifar10}
\centering
\resizebox{0.9\linewidth}{!}{
\begin{tabular}{cc}
\\
\hline
\toprule
(a) Generator & (b) Discriminator \\
\midrule
$z \in \mathbb{R}^{128} \sim \mathcal{N}(0, I)$ & image $x \in [-1, 1]^{32 \times 32 \times 3}$ \\
\midrule
dense,  $4\times 4\times 256$ & ResBlock down 128 \\
\midrule
ResBlock up 256 & ResBlock down 128 \\
\midrule
ResBlock up 256 & ResBlock down 128 \\
\midrule
ResBlock up 256 & ResBlock down 128 \\
\midrule
BN, ReLU, $3\times 3$ conv, Tanh & ReLU 0.1 \\
\midrule
& Global sum pooling \\
\midrule
& dense $\rightarrow$ 1 \\
\bottomrule
\end{tabular}
}
\end{minipage}%
\begin{minipage}{.5\linewidth}
\centering
\caption{
ResNet architecture for STL-10.}
\label{tab:arch-stl10}
\resizebox{0.9\linewidth}{!}{
\begin{tabular}{cc}
\\
\hline
\toprule
(a) Generator & (b) Discriminator \\
\midrule
$z \in \mathbb{R}^{128} \sim \mathcal{N}(0, I)$ & image $x \in [-1, 1]^{48 \times 48 \times 3}$ \\
\midrule
dense,  $6\times 6\times 512$ & ResBlock down 64 \\
\midrule
ResBlock up 256 & ResBlock down 128 \\
\midrule
ResBlock up 128 & ResBlock down 256 \\
\midrule
ResBlock up 64 & ResBlock down 512 \\
\midrule
BN, ReLU, $3\times 3$ conv, Tanh & ResBlock down 1024 \\
\midrule
& ReLU 0.1 \\
\midrule
& Global avg pooling \\
\midrule
& dense $\rightarrow$ 1 \\
\bottomrule

\end{tabular}
}
\end{minipage} 
\end{table}

\section{Algorithms}
\label{sect:appendix-algorithms}

In this section, we present the detailed algorithms for \val{DDA}, \valsub{ADAPT}{relax}, and \valsub{ADAPT}{strict}.

\subsection{Dynamic adjust algorithm}
\label{sect:appendix-algorithms-DA}

We first present \val{DDA} in \autoref{algo:DA}.
\begin{algorithm}
\caption{Dynamic density adjust (\val{DDA}) for the discriminator.}
\label{algo:DA}
\begin{algorithmic}[1]
{\small
\REQUIRE Generator $G$, discriminator $D$, BR upper bound $B_{+}$ and lower bound $B_{-}$, DA interval $\Delta T_D$, density increment $\Delta d$, current training iteration $t$. 
% \FOR{$t$ in $\left[1, \cdots, T\right]$}
\IF {$t\mod \Delta T_D==0$} 
    \STATE Compute time-averaged BR with \autoref{eq: BR}
    \IF {$\text{BR} > B_{+}$}
        \STATE Increase the density of discriminator from $d_D$ to $d_D + \Delta d$.
    \ELSIF{$\text{BR} < B_{-}$}
        \STATE Decrease the density of discriminator from $d_D$ to $d_D - \Delta d$.
    \ENDIF
    
\ENDIF
% \ENDFOR
}
\end{algorithmic}
\end{algorithm}

\begin{algorithm}
\caption{Relaxed balanced dynamic sparse training (\valsub{ADAPT}{relax}) for GANs.}
\label{algo:R-DDST}
\begin{algorithmic}[1]
{\small
\REQUIRE Generator $G$, discriminator $D$, total number of training iterations $T$, number of training steps for discriminator in each iteration $N$, discriminator adjustment interval $\Delta T_D$, DST interval for the generator $\Delta T_G$, density increment $\Delta d$, target generator density $d_G$, BR upper bound $B_{+}$ and lower bound $B_{-}$. 
\STATE Set initial discriminator density $d_D = d_G$
\FOR{$t$ in $\left[1, \cdots, T\right]$}
    
    \FOR{$n$ in $\left[1, \cdots, N\right]$}
        \STATE Compute the loss of discriminator $ \mathcal{L}_{D} (\bm{\theta}_{D})$ 
        \STATE  $ \mathcal{L}_{D} (\bm{\theta}_{D}).backward()$ 
    \ENDFOR
    
    \IF{$t\mod \Delta T_D==0$}
        \STATE Compute the loss of generator  $ \mathcal{L}_{G} (\bm{\theta}_{G})$ 
        \STATE $ \mathcal{L}_{G} (\bm{\theta}_{G}).backward()$ 
        \STATE Compute time-averaged BR with \autoref{eq: BR}
        \IF{ $\text{BR}>B_{+}$}
            \STATE Increase the density of discriminator from $d_D$ to $\min(100 \%,d_D + \Delta d)$.
        \ELSIF{$\text{BR}<B_{-}$}
            \STATE Decrease the density of discriminator from $d_D$ to $\max (0 \%, d_D - \Delta d)$.
        \ENDIF

    \ENDIF
    \IF{$t\mod \Delta T_G==0$}
        \STATE Apply DST to $G$
    \ENDIF
\ENDFOR
}
\end{algorithmic}
\end{algorithm}

\subsection{Relaxed balanced dynamic sparse training algorithm}
% \label{sect:appendix-algorithms2}
Details of \valsub{ADAPT}{relax} algorithm is presented in \autoref{algo:R-DDST}.

\subsection{Strict balanced dynamic sparse training algorithm}
% \label{sect:appendix-algorithms2}
Details of \valsub{ADAPT}{strict} algorithm is presented in \autoref{algo:S-DDST}.
\begin{algorithm}
\caption{Strict balanced dynamic sparse training (\valsub{ADAPT}{strict}) for GANs.}
\label{algo:S-DDST}
\begin{algorithmic}[1]
{\small
\REQUIRE Generator $G$, discriminator $D$, total number of training iterations $T$, number of training steps for discriminator in each iteration $N$, given maximal density of discriminator $d_\text{max}$, discriminator adjustment interval $\Delta T_D$, DST interval for the generator $\Delta T_G$, density increment $\Delta d$, target generator density $d_G$, BR upper bound $B_{+}$ and lower bound $B_{-}$. 
\STATE Set initial discriminator density $d_D = d_G$
\FOR{$t$ in $\left[1, \cdots, T\right]$}
    
    \FOR{$n$ in $\left[1, \cdots, N\right]$}
        \STATE Compute the loss of discriminator $ \mathcal{L}_{D} (\bm{\theta}_{D})$ 
        \STATE  $ \mathcal{L}_{D} (\bm{\theta}_{D}).backward()$ 
    \ENDFOR
    
    \IF{$t\mod \Delta T_D==0$}
        \STATE Compute the loss of generator  $ \mathcal{L}_{G} (\bm{\theta}_{G})$ 
        \STATE $ \mathcal{L}_{G} (\bm{\theta}_{G}).backward()$ 
        \STATE Compute time-averaged BR with \autoref{eq: BR}
        \IF{ $\text{BR}>B_{+}$ and $d_D<d_\text{max}$}
            \STATE Increase the density of discriminator from $d_D$ to $\min (d_\text{max}, d_D + \Delta d)$.
        \ELSIF{$\text{BR}>B_{+}$ and $d_D == d_\text{max}$}
            \STATE Apply DST to $D$
        \ELSIF{$\text{BR}<B_{-}$}
            \STATE Decrease the density of discriminator from $d_D$ to $\max (0 \%, d_D - \Delta d)$.
        \ENDIF

    \ENDIF
    \IF{$t\mod \Delta T_G==0$}
        \STATE Apply DST to $G$
    \ENDIF
\ENDFOR
}

\end{algorithmic}
\end{algorithm}

% \section{Balance ratio comparison of \val{STATIC} and \val{SDST}}
% \label{sect:SDST-STATIC-BR}
% In this section we show the BR comparison of \val{STATIC} and \val{SDST} in \autoref{sect:sdst}.

% \begin{figure*}[t]
%   \centering
%     \includegraphics[width=0.99\textwidth]{plots/STAT_SDST_comp_full.pdf}
%    \caption{BR comparison of \val{SDST} against \val{STATIC} sparse training for SNGAN on CIFAR-10 with different sparsity ratio combinations. The shaded areas denote the standard deviation.}
%    \label{fig:baseline12-BR}
%    \vspace{-0.5cm}
% \end{figure*}

% \textbf{\val{SDST} does not stabilize training.} As is shown in \autoref{fig:baseline12-BR}, \val{SDST} is unable to stabilize sparse GAN training when the training collapse occurs for \val{STATIC} methods. As is shown by first-row column 1; second-row column 1; third-row columns 1-2; fourth-row columns 1-3.

% \textbf{\val{SDST} increases the generator capacity.} However, when the discriminator is stronger than the generator where training collapse does not occur, \val{SDST} serves as a way to increase the generator capacity. As shown by higher BR values of \val{SDST} compared to \val{STATIC}. It suggests for each iteration, the generator is able to produce more realistic images to trick the discriminator.

\section{\val{ADAPT} balance ratio evolution}
\label{sect:DDST-BR}
In this section, we show that \val{ADAPT} methods are able to maintain a BR throughout training. We show the time evolution of BR and discriminator density for BigGAN on the CIFAR-10 dataset.

\begin{figure}[t]
  \centering
    \includegraphics[width=\textwidth]{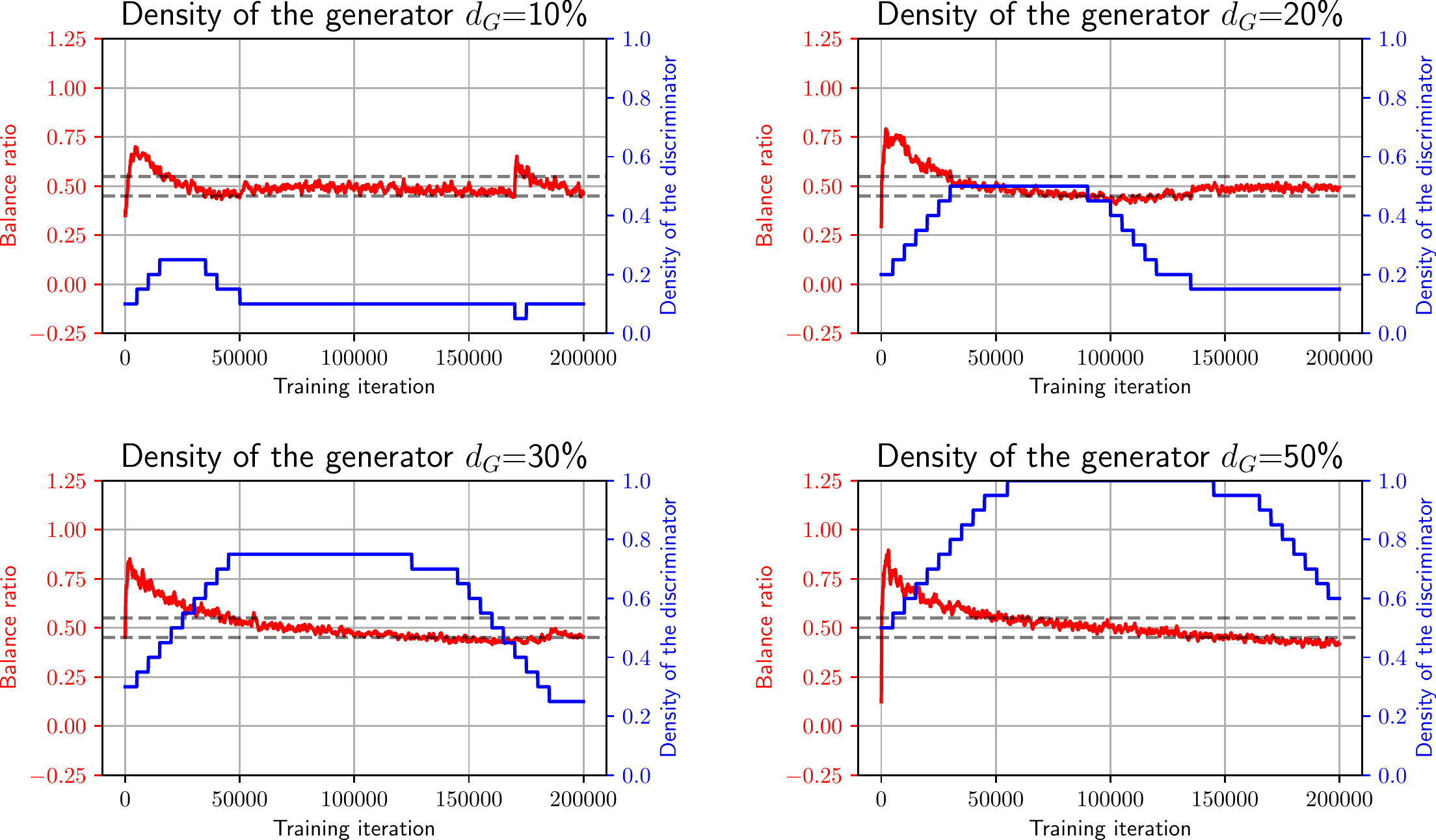} 
   \caption{Balance ratio and discriminator density evolution during training for \valsub{ADAPT}{relax} on BigGAN (CIFAR-10). Dashed lines represent BR values of 0.45 and 0.55.}
   \label{fig:RDDST-BR-CIFAR10} 
\end{figure}
\begin{figure}[t]
  \centering
    \includegraphics[width=\textwidth]{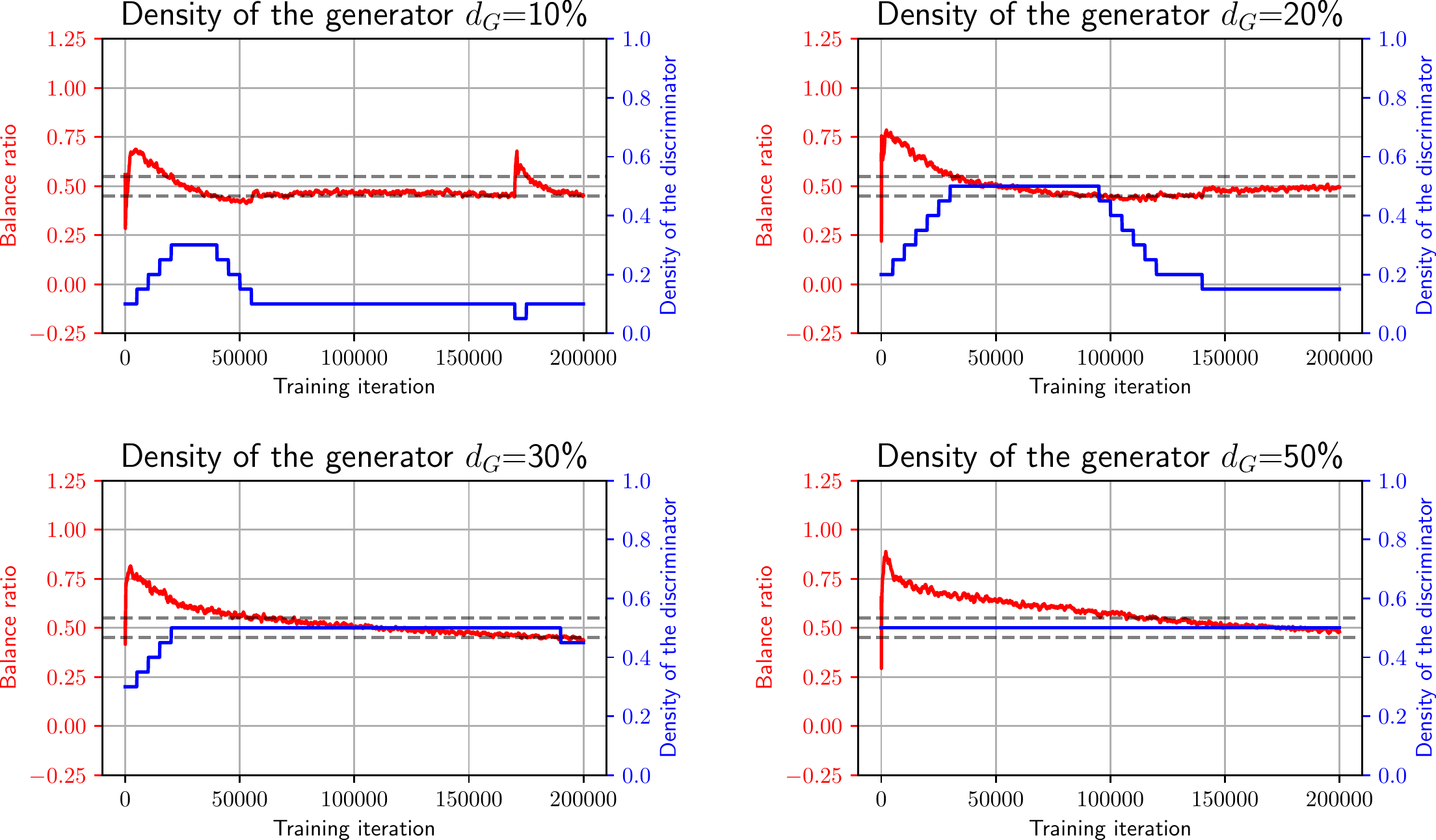} 
   \caption{Balance ratio and discriminator density evolution during training for \valsub{ADAPT}{strict} on BigGAN (CIFAR-10). Dashed lines represent BR values of 0.45 and 0.55.}
   \label{fig:SDDST-BR-CIFAR10}
\end{figure}

Results of \valsub{ADAPT}{relax} and \valsub{ADAPT}{strict} are shown in \autoref{fig:RDDST-BR-CIFAR10} and \autoref{fig:SDDST-BR-CIFAR10}. It clearly illustrates the ability of \val{ADAPT} to keep the BR controlled during GAN training.

\section{More experiment results}
\label{sect:appendix-more-results}
\subsection{IS and FID for the CIFAR-10 dataset}
In this section, we present corresponding IS scores results for \autoref{tab:benchmark-full} and \autoref{tab:benchmark-sparse}. The results are shown in \autoref{tab:benchmark-IS-full} and \autoref{tab:benchmark-IS-sparse}, respectively. We also include FID results of CIFAR-10 test set in \autoref{tab:benchmark-testCIFAR10-full}. 
\subsection{Naively applying DST to both the generator and the discriminator}
\label{sect:appendix-double_sdst}
In this section, we follow STU-GAN to compare the baseline where applying DST on both generators and discriminators. We name it \val{DST-bothGD}. 

We test on SNGAN (CIFAR-10) with $\Delta T_D = 1000$, $\Delta T_G=500$, and $\gamma=0.5$. Note that we use the balance strategy where $d_G=d_D$. The reason is that the strong strategy uses a dense discriminator, and it does not make sense to apply DST to a dense network.

We show the results in \autoref{tab:double-sdst}. It shows that it generates unstable results and consistenly performs worse than \val{SDST-Strong}. So we do not compare such baseline in the main body of the paper.

\subsection{Post-hoc pruning baseline}
In this section, we compare different sparse training methods with post-hoc magnitude pruning \citep{LTR} baseline. Magnitude pruning involves first training a dense generator, then pruning its weights globally based on their magnitudes. The pruned generator is then fine-tuned with the dense discriminator. We perform additional fine-tuning for $50\%$ of the original total iterations. Results are presented in \autoref{tab:posthoc}.

Our experimental results clearly demonstrate the advantages of dynamic sparse training over post-hoc magnitude pruning. The latter typically requires around 150\% normalized training FLOPs, while DST methods constantly achieve comparable or better performance with significantly reduced computational cost.

\begin{table}
    \caption{FID ($\downarrow$) of different sparse training methods along with post-hoc pruning baseline \textbf{with no constraint on the density of the discriminator}. Best results are in \textbf{bold}; second-best results are \underline{underlined}.}
    \label{tab:posthoc}
    \resizebox{\columnwidth}{!}{
    \begin{tabular}{ l cccc cccc cccc}
    \hline
    \toprule
    Dataset  &\multicolumn{4}{c}{\textbf{CIFAR-10 (SNGAN)}} &\multicolumn{4}{c}{\textbf{STL-10 (SNGAN)}} &\multicolumn{4}{c}{\textbf{CIFAR-10 (BigGAN)}} \\
    \midrule
    Generator density & 10\% & 20\%  & 30\% & 50\% & 10\% & 20\%  & 30\% & 50\%  & 10\% & 20\%  & 30\% & 50\%\\
    \midrule
    (Dense Baseline)   & \multicolumn{4}{c}{10.74} & \multicolumn{4}{c}{29.71}  & \multicolumn{4}{c}{8.11}  \\
    \midrule
    Post-hoc pruning & 20.89 & 14.07 & 12.99 & 11.90 & 57.28 & 37.12 & 31.98 & \textbf{29.70} & 15.44 & 10.84 & 9.65 & 8.77 \\
    \midrule
    STATIC-Balance  & 26.75 & 19.04 & 15.05 & 12.24 & 48.18 & 44.67 & 41.73 & 37.68 & 16.98 & 12.81 & 10.33 & 8.47 \\
    STATIC-Strong  & 26.79& 19.65 & 14.38 & 11.91 & 52.48 & 43.85 & 42.06 & 37.47& 23.48 & 14.26 & 11.19 & 8.64  \\
    \midrule
    \midrule
    \tikzdiamond[blue, fill=blue] SDST-Balance-SET  & 26.23 & 17.79 & 13.21 & 11.79 & 56.41 & 46.58 & 39.93 & 30.37 & 12.41 & 9.87 & 9.13 & \textbf{8.01}\\
    \tikzdiamond[blue, fill=blue] SDST-Strong-SET  & \underline{16.49} & \underline{13.36} & \textbf{11.68} & \underline{10.68} & 67.37 & 49.96 & 37.99 & 31.08& 18.94 & 9.64 & 8.75 & 8.36 \\

    \tikzcircle[red,fill=red] SDST-Balance-RigL  &  27.06 & 16.36 & 14.00 & 12.28 & \underline{43.08} & 33.90 & 31.83 & 30.30 & 12.45 & 9.42 & 8.86 & \underline{8.03} \\
    \tikzcircle[red,fill=red] SDST-Strong-RigL  & 17.02 & 13.86 & 12.51 & 11.35 & 53.65 & \underline{33.25} & \textbf{31.41} & 30.18& \underline{10.58} & \underline{9.11} & \underline{8.69} & 8.33 \\

    \midrule
    \midrule
    $\text{ADAPT}_\text{relax}$ (Ours)  & \textbf{14.19} & \textbf{13.19} & \underline{12.38} & \textbf{10.60} & \textbf{35.98} & \textbf{33.06} & \underline{31.71} & \underline{29.96}& \textbf{10.19} & \textbf{8.56} & \textbf{8.36} & 8.22 \\
    
    \bottomrule

    \end{tabular}
    }
    \vspace{-0.5cm}
\end{table}

\begin{table}
    \centering
    \caption{
    FID ($\downarrow$) of different sparse training methods on CIFAR-10 datasets with no constraint on the density of the discriminator. Best results are in \textbf{bold}; second-best results are \underline{underlined}.}
    \label{tab:double-sdst}
    \resizebox{0.5 \columnwidth}{!}{
    \begin{tabular}{ l ccccccc}
    \\
    \hline
    \toprule
    Dataset  &\multicolumn{4}{c}{\textbf{CIFAR-10}}  \\
    \midrule
    Generator density & 10\% & 20 \%  & 30 \% & 50 \%  \\
    \midrule
    (Dense Baseline)   & \multicolumn{4}{c}{10.74}  \\
    \midrule
    Static-Balance  & 26.75 & 19.04 & 15.05 & 12.24 \\
    Static-Strong  & 26.79& 19.65 & 14.38 & 11.91 \\
    \midrule
    \midrule
    \tikzdiamond[blue, fill=blue] DST-bothGD-SET & 20.57 & 14.90 & 12.58 & 11.86\\
    \tikzcircle[red,fill=red] DST-bothGD-RigL & 31.95 & 17.99 & 13.24 & 12.47 \\
    \midrule
    \midrule
    \tikzdiamond[blue, fill=blue] SDST-Balance-SET  & 26.23 & 17.79 & 13.21 & 11.79  \\
    \tikzdiamond[blue, fill=blue] SDST-Strong-SET  & \underline{16.49} & \underline{13.36} & \textbf{11.68} & \underline{10.68} \\

    \tikzcircle[red,fill=red] SDST-Balance-RigL  &  27.06 & 16.36 & 14.00 & 12.28  \\
    \tikzcircle[red,fill=red] SDST-Strong-RigL  & 17.02 & 13.86 & 12.51 & 11.35 \\

    \midrule
    \midrule
    % \tikzcircle[red,fill=red] $\text{ADAPT}_\text{relax}$ (RigL)  & \textbf{14.19} & \textbf{13.19} & \underline{12.38} & \textbf{10.60} & \textbf{35.98} & \textbf{33.06} & \underline{31.71} & \textbf{29.96}\\
    $\text{ADAPT}_\text{relax}$ (Ours)  & \textbf{14.19} & \textbf{13.19} & \underline{12.38} & \textbf{10.60} \\
    
    \bottomrule

    \end{tabular}
    }
    \vspace{-1mm}
\end{table}

\begin{table}
    \centering
    \caption{
    IS (higher is better) of different sparse training methods. There is no constraint on the density of the discriminator, i.e., $d_\text{max}=100\%$.}
    \label{tab:benchmark-IS-full}
    \resizebox{0.99\linewidth}{!}{
    \begin{tabular}{ l cccc cccc cccc cccc}
    \\
    \hline
    \toprule
    Dataset & \multicolumn{4}{c}{\textbf{SNGAN(CIFAR-10)}} &\multicolumn{4}{c}{\textbf{SNGAN(STL-10)}} &\multicolumn{4}{c}{\textbf{BigGAN(CIFAR-10)}} &\multicolumn{4}{c}{\textbf{BigGAN(TinyImageNet)}}\\
    \midrule
    Generator density & 10\% & 20 \%  & 30 \% & 50 \% & 10\% & 20 \%  & 30 \% & 50 \% & 10\% & 20 \%  & 30 \% & 50 \% & 10\% & 20 \%  & 30 \% & 50 \% \\
    \midrule
    (Dense Baseline) & \multicolumn{4}{c}{8.48} & \multicolumn{4}{c}{9.16} & \multicolumn{4}{c}{8.99} & \multicolumn{4}{c}{14.65}\\
    \midrule
    Static-Balance & 7.24 & 7.83 & 8.06 & 8.38 & 7.94 & 8.19 & 8.44 & 8.69 & 7.99 & 8.24 & 8.68 & 8.90 & 10.65 & 12.28 & 13.41 & 13.57\\
    Static-Strong & 7.52 & 8.03 & 8.32 & 8.45 & 7.70 & 8.22 & 8.35 & 8.70 & 7.75 & 8.13 & 8.52 & 8.99 & 10.45 & 12.56 & 13.61 & 13.73\\
    \midrule

    \tikzdiamond[blue, fill=blue] SDST-Balance-SET & 7.28 & 7.89 & 8.22 & 8.38 & 8.43 & 8.92 & 9.26 & 9.31 & 8.62 & 8.67 & 8.82 & 8.98 & 11.75 & 12.60 & 12.30 & 12.21 \\
    \tikzdiamond[blue, fill=blue] SDST-Strong-SET & 8.37 & 8.54 & 8.57 & 8.60 & 7.65 & 8.53 & 9.39 & 9.21 & 8.16 & 8.78 & 8.85 & 9.06 & 12.75 & 12.84 & 12.46 & 13.73\\
    \tikzcircle[red,fill=red] SDST-Balance-RigL & 7.19 & 7.94 & 8.18 & 8.34 & 8.98 & 9.07 & 9.12 & 9.28 & 8.64 & 8.71 & 8.91 & 8.93 & 12.67 & 13.32 & 13.18 & 13.61\\
    \tikzcircle[red,fill=red] SDST-Strong-RigL & 8.32 & 8.52 & 8.59 & 8.57 & 8.15 & 9.10 & 9.16 & 9.17 & 8.65 & 8.72 & 8.97 & 9.00 & 13.32 & 13.35 & 13.60 & 14.47 \\
    \midrule
    \midrule

    ADAPT$_\text{relax}$ (Ours) & 8.42 & 8.44 & 8.54 & 8.60 & 9.08 & 9.29 & 9.06 & 9.26 & 8.74 & 9.07 & 8.98 & 9.00 & 13.09 & 13.57 & 13.68 & 15.77\\
    \bottomrule

    \end{tabular}
    }
    \caption{
    IS (higher is better) of different sparse training methods. The density of the discriminator is constrained to be lower than $d_\text{max}=50\%$.}
    \label{tab:benchmark-IS-sparse}
    \resizebox{0.99\linewidth}{!}{
    \begin{tabular}{ l cccc cccc cccc cccc}
    \\
    \hline
    \toprule
    Dataset & \multicolumn{4}{c}{\textbf{SNGAN(CIFAR-10)}} &\multicolumn{4}{c}{\textbf{SNGAN(STL-10)}} &\multicolumn{4}{c}{\textbf{BigGAN(CIFAR-10)}} &\multicolumn{4}{c}{\textbf{BigGAN(TinyImageNet)}}\\
    \midrule
    Generator density & 10\% & 20 \%  & 30 \% & 50 \% & 10\% & 20 \%  & 30 \% & 50 \% & 10\% & 20 \%  & 30 \% & 50 \% & 10\% & 20 \%  & 30 \% & 50 \% \\
    \midrule
    (Dense Baseline) & \multicolumn{4}{c}{8.48} & \multicolumn{4}{c}{9.16} & \multicolumn{4}{c}{8.99} & \multicolumn{4}{c}{14.65}\\
    \midrule
    Static-Balance & 7.24 & 7.83 & 8.06 & 8.38 & 7.94 & 8.19 & 8.44 & 8.69 & 7.99 & 8.24 & 8.68 & 8.90 & 10.65 & 12.28 & 13.41 & 13.57\\
    Static-Strong & 7.85 & 8.14 & 8.31 & 8.38 & 7.89 & 8.22 & 8.38 & 8.69 & 7.75 & 8.03 & 8.52 & 8.90 &  9.99 & 11.61 & 13.77 & 13.57\\
    \midrule

    \tikzdiamond[blue, fill=blue] SDST-Balance-SET & 7.28 & 7.89 & 8.22 & 8.38 & 8.43 & 8.92 & 9.26 & 9.31 & 8.62 & 8.67 & 8.82 & 8.98 & 11.75 & 12.60 & 12.30 & 12.21\\
    \tikzdiamond[blue, fill=blue] SDST-Strong-SET & 8.33 & 8.53 & 8.40 & 8.38 & 8.50 & 8.77 & 9.46 & 9.26 & 8.55 & 8.77 & 8.84 & 8.98 &12.00 & 12.87 & 12.16 & 12.21\\
    \tikzcircle[red,fill=red] SDST-Balance-RigL & 7.19 & 7.94 & 8.18 & 8.34 & 8.98 & 9.07 & 9.12 & 9.28 & 8.64 & 8.71 & 8.91 & 8.93 & 12.67 & 13.32 & 13.18 & 13.61\\
    \tikzcircle[red,fill=red] SDST-Strong-RigL &  8.24 & 8.48 & 8.37 & 8.34 & 8.28 & 9.05 & 9.11 & 9.28 & 8.61 & 8.83 & 8.84 & 8.93 & 12.04 & 12.66 & 13.57 & 13.61\\
    \midrule
    \midrule

    ADAPT$_\text{strict}$ (Ours) & 8.27 & 8.36 & 8.48 & 8.47 &  8.98 & 9.17 & 9.20 & 9.19 & 8.90 & 8.89 & 8.92 & 9.10 & 13.85 & 13.61 & 14.05 & 14.40\\
    \bottomrule

    \end{tabular}
    }
\end{table}

\begin{table}
    \centering
    \caption{
    FID of test set ($\downarrow$) of different sparse training methods on SNGAN (CIFAR-10) dataset. Best results are in \textbf{bold}; second-best results are \underline{underlined}.}
    \label{tab:benchmark-testCIFAR10-full}
    \resizebox{0.95\linewidth}{!}{
    \begin{tabular}{ l cccccccc}
    \\
    \hline
    \toprule
    Maximal discriminator density $d_\text{max}$ & \multicolumn{4}{c}{\textbf{100 \%}} & \multicolumn{4}{c}{\textbf{50 \%}}\\
    \midrule
    Generator density & 10\% & 20 \%  & 30 \% & 50 \% & 10\% & 20 \%  & 30 \% & 50 \% \\
    % \midrule
    % Sparsity percentage & 89.26\% & 95.60\% &  89.26\% & 95.60\% \\ 
    \midrule
    (Dense Baseline) & \multicolumn{8}{c}{13.32} \\
    \midrule
    Static-Balance & 29.56 & 21.79 & 17.80 & 14.94 & 29.56 & 21.79 & 17.80 & 14.94\\
    Static-Strong & 29.50 & 22.45 & 17.12 & 14.58 & 24.62 & 19.43 & 16.32 & 14.94\\
    \midrule

    \tikzdiamond[blue, fill=blue] SDST-Balance-SET & 28.84 & 20.31 & 15.95 & 14.35  & 28.84 & 20.31 & 15.95 & \textbf{14.35}\\
    \tikzdiamond[blue, fill=blue] SDST-Strong-SET & \underline{19.16} & \underline{16.12} & \textbf{14.45} & \underline{13.50} & 18.38 & \textbf{15.33} & \textbf{14.78} & \textbf{14.35}\\
    \tikzcircle[red,fill=red] SDST-Balance-RigL & 29.77 & 19.02 & 16.68 & 15.05 & 29.77 & 19.02 & 16.68 & 15.05 \\
    \tikzcircle[red,fill=red] SDST-Strong-RigL & 19.72 & 16.50 & 15.20 & 14.09 & \underline{17.92} & \underline{15.51} & 15.52 & 15.05 \\
    \midrule
    \midrule
    % L-DDST-RigL & \underline{13.12} & 14.71 & 12.74 & 11.54 & 42.64 & 31.47 &32.11 & 30.76\\

    ADAPT$_\text{relax}$ (Ours) &\textbf{16.82}&\textbf{15.85}&\underline{15.14} & \textbf{13.37}&-&-&-&- \\ 
    ADAPT$_\text{strict}$ (Ours) &-&-&-&- &\textbf{17.19}& 15.57& \underline{14.92}& \underline{14.80}\\
    
    \bottomrule

    \end{tabular}
                                                        }
\end{table}

\begin{table}
    \centering
    \caption{
    FID of test set ($\downarrow$) of different sparse training methods on BigGAN (CIFAR-10) dataset. Best results are in \textbf{bold}; second-best results are \underline{underlined}.}
    \label{tab:benchmark-testCIFAR10-biggan-full}
    \resizebox{0.95\linewidth}{!}{
    \begin{tabular}{ l cccccccc}
    \\
    \hline
    \toprule
    Maximal discriminator density $d_\text{max}$ & \multicolumn{4}{c}{\textbf{100 \%}} & \multicolumn{4}{c}{\textbf{50 \%}}\\
    \midrule
    Generator density & 10\% & 20 \%  & 30 \% & 50 \% & 10\% & 20 \%  & 30 \% & 50 \% \\
    % \midrule
    % Sparsity percentage & 89.26\% & 95.60\% &  89.26\% & 95.60\% \\ 
    \midrule
    (Dense Baseline) & \multicolumn{8}{c}{10.36} \\
    \midrule
    Static-Balance & 19.58 & 15.63 & 13.21 & 10.92 & 19.58 & 15.63 & 13.21 & 10.92 \\
    Static-Strong & 26.08 & 15.82 & 13.47 & 10.95 & 22.04 & 16.39 & 13.73 & 10.92\\
    \midrule

    \tikzdiamond[blue, fill=blue] SDST-Balance-SET & 14.90 & 12.77 & 11.82 & \textbf{10.68} & 14.90 & 12.77 & 11.82 & \underline{10.68}\\
    \tikzdiamond[blue, fill=blue] SDST-Strong-SET & 21.63 & 11.92 & 11.27 & \underline{10.75} & 14.53 & \underline{11.83} & 10.96 & \underline{10.68}\\
    \tikzcircle[red,fill=red] SDST-Balance-RigL & 14.86 & 12.03 & 11.30 & \textbf{10.68}  & 14.86 & 12.03 & 11.30 & \underline{10.68}\\
    \tikzcircle[red,fill=red] SDST-Strong-RigL & \underline{13.35} & \underline{11.58} & \underline{11.00} & 10.88 & \underline{12.59} & 12.03 & \textbf{10.89} & \underline{10.68} \\
    \midrule
    \midrule
    % L-DDST-RigL & \underline{13.12} & 14.71 & 12.74 & 11.54 & 42.64 & 31.47 &32.11 & 30.76\\

    ADAPT$_\text{relax}$ (Ours) &\textbf{12.71} & \textbf{11.02} & \textbf{10.62} & 10.80 &-&-&-&- \\ 
    ADAPT$_\text{strict}$ (Ours) &-&-&-&- & \textbf{11.83} & \textbf{11.22} & \underline{10.92} & \textbf{10.33}\\
    \bottomrule

    \end{tabular}
                                                        }
\end{table}

\section{A detailed comparison of training costs}
\label{sect:appendix-flops-accurate}
In this section, we include the detailed computational cost of all sparse training methods. More specifically, we take into account the density redistribution over different layers in this section. Also, we make an assumption that the computational overhead introduced by computing BR can be neglected.\footnote{This is true if we compute BR less frequently.}

Here we provide training costs for the \textbf{strict} setting in \autoref{tab:benchmark-sparse-flops}.

\begin{table}
    \caption{Normalized training FLOPs ($\downarrow$) of different sparse training methods. \textbf{The density of the discriminator is constrained to be lower than 50\%}.}
    \label{tab:benchmark-sparse-flops}
    \resizebox{\columnwidth}{!}{
    \begin{tabular}{ l cccc cccc cccc cccc}
    \hline
    \toprule
    Dataset  &\multicolumn{4}{c}{\textbf{CIFAR-10 (SNGAN)}} &\multicolumn{4}{c}{\textbf{STL-10 (SNGAN)}} &\multicolumn{4}{c}{\textbf{CIFAR-10 (BigGAN)}} &\multicolumn{4}{c}{\textbf{TinyImageNet (BigGAN)}}\\
    \midrule
    Generator density & 10\% & 20\%  & 30\% & 50\% & 10\% & 20\%  & 30\% & 50\%  & 10\% & 20\%  & 30\% & 50\% & 10\% & 20\%  & 30\% & 50\%\\
    \midrule
    (Dense Baseline)   & \multicolumn{4}{c}{100\% ($2.67 \times 10^{17}$)} & \multicolumn{4}{c}{100\% ($3.94 \times 10^{17}$)}  & \multicolumn{4}{c}{100\% ($6.81 \times 10^{17}$)}  & \multicolumn{4}{c}{100\% ($9.85 \times 10^{17}$)} \\
    \midrule
    Static-Balance   
    & 8.97\% & 17.08\% & 26.25\% & 47.25\%   
    & 27.30\% & 47.14\% & 59.22\% & 73.35\%
    & 9.79\% & 19.02\% & 28.66\% & 49.03\%
    & 23.25\% & 44.87\% & 60.91\% & 79.29\%
    \\
    Static-Strong  
& 30.89\% & 33.58\% & 37.17\% & 47.25\%
& 70.65\% & 71.48\% & 72.14\% & 73.35\%
&  42.66\% & 43.69\% & 45.10\% & 49.03\%
&  41.52\% & 55.03\% & 66.29\% & 79.29\%
    \\
    \midrule
    \midrule
    \tikzdiamond[blue, fill=blue] SDST-Balance-SET  
& 9.78\% & 18.91\% & 28.35\% & 48.44\%
& 27.55\% & 47.60\% & 60.17\% & 75.38\%
& 10.35\% & 20.12\% & 29.96\% & 49.82\%
& 21.13\%& 37.06\%&48.83\%&65.58\%
    \\

    \tikzdiamond[blue, fill=blue] SDST-Strong-SET  
& 31.87\% & 35.51\% & 39.53\% & 48.44\%
& 70.95\% & 71.97\% & 73.07\% & 75.38\%
&43.25\% & 44.80\% & 46.42\% & 49.82\%
& 39.28\% & 47.31\% & 54.11\% & 65.58\%
    \\

    \tikzcircle[red,fill=red] SDST-Balance-RigL  
& 10.71\% & 17.43\% & 25.66\% & 43.56\%
& 29.51\% & 50.41\% & 63.34\% & 79.03\%
& 9.92\% & 19.30\% & 28.90\% & 48.31\%
& 24.97\% & 43.86\% & 57.26\% & 76.75\%
    
    \\
    \tikzcircle[red,fill=red] SDST-Strong-RigL  
& 31.22\% & 33.93\% & 36.63\% & 43.56\%
& 72.95\% & 75.05\% & 76.42\% & 79.03\%
& 42.80\% & 44.08\% & 45.37\% & 48.31\%
& 43.76\% & 53.71\% & 63.05\% & 76.75\%
    \\

    \midrule
    \midrule
    $\text{ADAPT}_\text{strict}$ (Ours) 
& 24.23\% & 27.55\% & 31.70\% & 37.83\%
& 50.91\% & 70.18\% & 75.99\% & 80.68\%
& 10.32\% & 23.69\% & 31.54\% & 33.83\%
& 34.42\% & 51.68\% & 62.34\% & 77.46\%
    \\
    
    \bottomrule

    \end{tabular}
    }
    \vspace{-0.5cm}
\end{table}

% \subsection{SNGAN on the CIFAR-10 dataset}
% We first show the results of SNGAN (CIFAR-10) in \autoref{tab:appendix-flops-sncifar10-relax} and \autoref{tab:appendix-flops-sncifar10-strict}. The results show that generally \val{ADAPT} is able to achieve promising performance with reasonable computational costs. 

% \input{table/FLOPs/table_SNCIFAR10_relax.tex}
% \input{table/FLOPs/table_SNCIFAR10_strict.tex}

% \subsection{SNGAN on the STL-10 dataset}
% We then show the results of SNGAN (STL-10) in \autoref{tab:appendix-flops-snstl10-relax} and \autoref{tab:appendix-flops-snstl10-strict}. The results show that generally \val{ADAPT} is able to achieve promising performance with reasonable computational costs. 

% \input{table/FLOPs/table_SNSTL10_relax.tex}
% \input{table/FLOPs/table_SNSTL10_strict.tex}

\end{document}